\documentclass{article}

    \PassOptionsToPackage{numbers, compress}{natbib}
\usepackage[preprint]{neurips_2026}

\usepackage[utf8]{inputenc} % allow utf-8 input
\usepackage[T1]{fontenc}    % use 8-bit T1 fonts
\usepackage{hyperref}       % hyperlinks
\usepackage{url}            % simple URL typesetting
\usepackage{booktabs}       % professional-quality tables
\usepackage{amsfonts}       % blackboard math symbols
\usepackage{nicefrac}       % compact symbols for 1/2, etc.
\usepackage{microtype}      % microtypography
\usepackage{xcolor}         % colors
\usepackage{makecell}
\usepackage{colortbl}
\usepackage{tabularx}
\usepackage{amsmath}
\usepackage{caption}
\usepackage{booktabs}
\usepackage{graphicx}
\usepackage{subcaption}
\usepackage{pifont}
\usepackage{threeparttable}
\usepackage{placeins}

\title{VL-DINO: Leveraging CLIP Vision-Language Knowledge for Open-Vocabulary Object Detection}

\author{%
  Hao Zhang\textsuperscript{1} \quad
  Qinran Lin\textsuperscript{1} \quad
  Linqi Song\textsuperscript{2} \quad
  Yong Li\textsuperscript{1} \thanks{Corresponding author.} \\
  \textsuperscript{1}Chongqing University, Chongqing, China \\
  \textsuperscript{2}City University of Hong Kong, Hong Kong, China \\
  \texttt{zhangh@stu.cqu.edu.cn} \\ \texttt{yongli@cqu.edu.cn} 
}

\begin{document}

\maketitle

\begin{abstract}
  Vision-language models like CLIP can provide rich semantic priors for open-vocabulary object detection. However, jointly integrating both textual and visual knowledge into detection architectures remains challenging. In this paper, we propose \textbf{VL-DINO}, an open-vocabulary detector that enhances DINO through more effective exploitation of CLIP's \textbf{V}\textit{ision}-\textbf{L}\textit{anguage} knowledge. Specifically, a \textbf{Q}\textit{uery-guided} \textbf{P}\textit{ositive} \textbf{S}\textit{ample} \textbf{C}\textit{onstruction} (QPSC) module is first developed to construct additional high-quality positive samples, enabling the vanilla DINO framework to better accommodate mixed training across heterogeneous data sources while providing more vision-language alignment signals, thereby incorporating richer textual knowledge during training. A \textbf{V}\textit{isual} \textbf{S}\textit{emantic} \textbf{E}\textit{ncoder} (VSE) module is then introduced to distill CLIP visual knowledge into backbone-extracted features, producing fused features for subsequent encoder refinement. Based on the fused features, an \textbf{O}\textit{bject}-\textbf{R}\textit{egion} \textbf{S}\textit{emantic} \textbf{A}\textit{lignment} (ORSA) module extracts object-centric region features and aligns them with the corresponding textual embeddings, further incorporating textual cues. In the zero-shot setting, VL-DINO-T and VL-DINO-L achieve 36.3 and 38.1 AP on LVIS benchmark, respectively, consistently outperforming prior advanced approaches. Extensive experiments demonstrate the effectiveness and competitive performance of the proposed design. The code and model are available at \url{https://github.com/HaoZ416/VL-DINO}.
\end{abstract}

\section{Introduction}
\label{intro}

Open-vocabulary detection(OVD) aims to overcome the inherent limitation of conventional object detectors, whose predictions at inference time are restricted to a predefined set of categories~\cite{focal-loss31,fast-rcnn47,yolo48}. Recent OVD approaches endow detectors with open-vocabulary generalization by incorporating the textual semantic knowledge of vision-language models(VLM)~\cite{Radford1, Jia2} into traditional detection frameworks to meet the demands of real-world applications~\cite{yolo-world3,yoloe4,T-rex2-5,gdino6,glip7}. Beyond transferring textual semantic knowledge, existing methods have also explored how to leverage the visual semantic knowledge encoded in the CLIP~\cite{Radford1} image encoder. These efforts mainly leverage the image encoder as a teacher to guide the detector, distilling its rich visual knowledge into refined region-level features through feature alignment or supervision signals~\cite{baron8,vild9,detpro10,oadp11}.

However, existing approaches still suffer from several limitations. First, methods for transferring textual semantic knowledge typically depend on layer-wise cross-modal fusion modules, resulting in increased model complexity~\cite{gdino6,glip7,yolo-world3}. Second, methods for exploiting visual semantic knowledge are fundamentally constrained by the pretraining nature of CLIP: trained with coarse-grained image-text supervision, CLIP mainly learns globally aligned visual representations, and its dense visual features are therefore suboptimal~\cite{maskclip13,sclip14,naclip15,clearclip16}, which weakens the effectiveness of direct feature distillation into detectors. Finally, the joint integration of textual and visual semantic knowledge into a unified detection framework has not been fully explored.

To overcome the above limitations, we propose VL-DINO, an open-vocabulary detector built upon DINO~\cite{dino17} and CLIP. A key obstacle in adapting DINO to OVD lies in its original denoising design, which requires the total number of unique dataset-specific categories to be known in advance. This assumption does not hold in the OVD setting, where grounding datasets~\cite{flickr30k18,gqa19} are often used to train models, containing a large number of free-form phrases (e.g., ``a brown building'') that cannot be enumerated with predefined category indices. Consequently, the application of the original denoising mechanism in DINO becomes infeasible, and the model mainly relies on vanilla object queries to align with textual semantics. In light of this, we develop a \textbf{Q}\textit{uery-guided} \textbf{P}\textit{ositive} \textbf{S}\textit{ample} \textbf{C}\textit{onstruction} module (QPSC), which constructs additional high-quality positive samples without relying on category-specific labels. QPSC introduces no extra learnable parameters, and all positive samples preserve the denoising mechanism advantage of bypassing bipartite matching while directly contributing to classification and box regression, and meanwhile provides richer vision-language alignment signals, allowing the detector to acquire richer textual semantic knowledge.

Additionally, as discussed above, CLIP mainly captures global semantics, making its dense local features less effective for direct region-level distillation. Motivated by this observation, we introduce a \textbf{V}\textit{isual} \textbf{S}\textit{emantic} \textbf{E}\textit{ncoder} (VSE), which takes the high-level features extracted by the detector backbone(e.g., Swin Transformer~\cite{swin20}) as input and distill CLIP visual semantics into corresponding feature outputs. This enriches the high-level features with stronger global semantic information, while keeping the low-level detail-rich features unchanged. The distilled high-level features are then fused with their corresponding original backbone features, and together with the unchanged low-level features, are fed into the encoder for further refinement of dense local information.

Finally, we introduce an \textbf{O}\textit{bject}-\textbf{R}\textit{egion} \textbf{S}\textit{emantic} \textbf{A}\textit{lignment} (ORSA) module, which aligns foreground object-region features of the first encoder layer with their corresponding textual semantics. By doing so, ORSA encourages fused features in the shallow encoder layers to be more sensitive to textual semantic cues, making them better adapted to OVD task while further injecting textual semantic knowledge into the detector.

Overall, these modules enable VL-DINO to effectively absorb the large-scale semantic priors of CLIP from both textual and visual perspectives, rather than relying on a single modality alone. Notably, QPSC and ORSA are used only during training and are discarded at inference time. As a result, beyond DINO detector and CLIP model, only VSE module is retained as an additional component, preserving the overall simplicity of the architecture. Extensive experiments in the zero-shot setting on LVIS~\cite{lvis} and COCO~\cite{COCO} benchmarks demonstrate the effectiveness of the proposed design. VL-DINO-T and VL-DINO-L achieve 36.3 and 38.1 AP on LVIS, and 48.5 and 50.0 AP on COCO, respectively. Our contributions are summarized as follows:
\begin{itemize}
    \item We propose QPSC, which constructs additional high-quality positive samples to provide DINO with richer textual semantic supervision.
    \item We introduce VSE, which distills CLIP visual semantic knowledge into high-level backbone features and enhance their global semantic representations.
    \item We present ORSA, which aligns shallow object-region features with textual embeddings to further integrate textual semantic knowledge into the detector.
    \item Extensive zero-shot experiments on LVIS and COCO verify the effectiveness of VL-DINO and demonstrate its strong generalizability and competitive performance.
\end{itemize}

\section{Related work}
\label{related work}

\noindent\textbf{DETR-like Models.} DETR~\cite{detr27} introduces Transformer~\cite{tfer21} into object detection and establishes a fully end-to-end detection framework. Building upon this paradigm, a series of follow-up works have focused on improving its optimization and representation learning. Deformable DETR~\cite{deformable-detr28} extends DETR with deformable attention, enabling more effective modeling of multi-scale visual features in the encoder. To alleviate the training instability caused by Hungarian matching, subsequent methods~\cite{dn-detr29,dino17} incorporate denoising mechanisms. Meanwhile, although the one-to-one matching scheme in DETR eliminates hand-crafted designs such as non-maximum suppression and anchor boxes, it also results in sparse supervision. To address this limitation, later works~\cite{group-detr23,hdetr24,co-detr25,deim26} draw inspiration from conventional CNN-based detectors~\cite{focal-loss31,fcos32} and introduce one-to-many training branches, thereby providing denser supervision signals and improving feature representation.

\noindent\textbf{Open-Vocabulary Detectors.} The rise of large-scale pretrained vision-language models~\cite{Radford1, Jia2} has made it practical for OVD detectors to exploit large amounts of vision-language knowledge~\cite{T-rex2-5,regionclip38,39,detclip40,dinox41}. A straightforward line of research transfers such semantic knowledge through knowledge distillation~\cite{baron8,vild9,detpro10,oadp11}. Another representative direction is to unify detection and grounding, allowing the model to benefit from both types of supervision. In this vein, GLIP~\cite{glip7} is among the first to cast detection as grounding and jointly train on detection and grounding datasets, while GroundingDINO~\cite{gdino6} further strengthens multimodal understanding through layer-wise cross-modal fusion and decoding. More recently, YOLO-World~\cite{yolo-world3}and YOLOE~\cite{yoloe4} push OVD toward lightweight and real-time regimes, improving its suitability for deployment on edge devices.

\begin{figure}
  \centering
  \includegraphics[width=\textwidth]{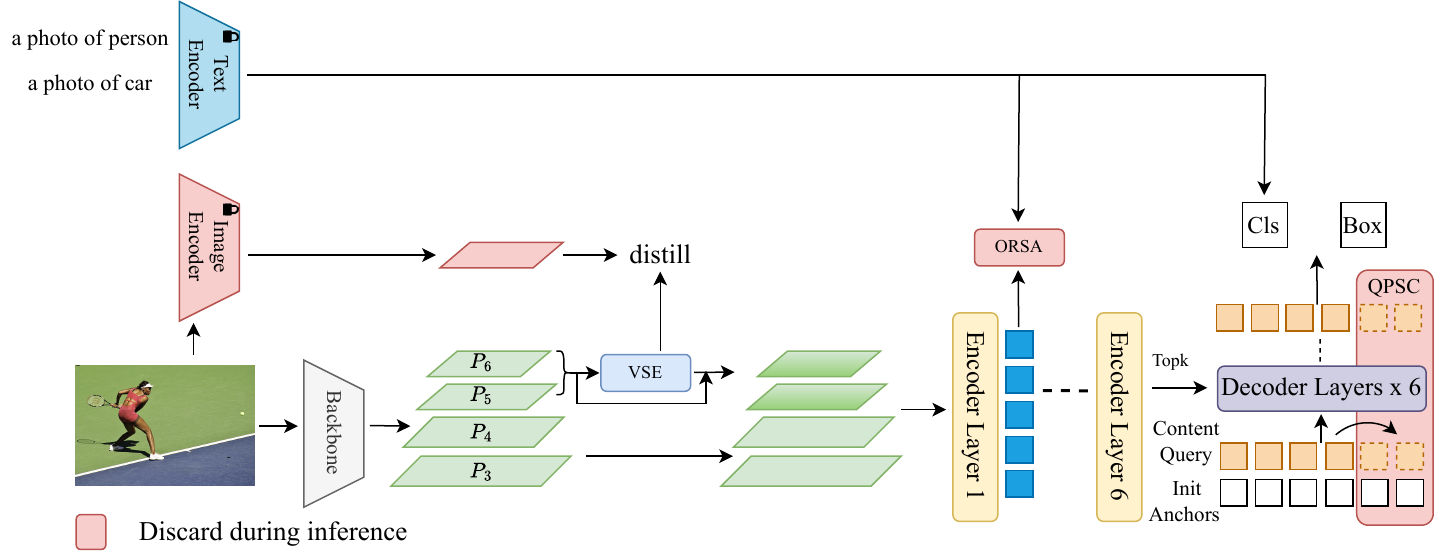}
  \caption{Overview of the VL-DINO model. During training, QPSC leverages reliable content queries to provide additional high-quality positive samples, thereby offering richer textual supervision for vision--language alignment. Given an input image $I$, the distillation branch employs a CLIP image encoder to extract visual semantic knowledge and distills it into the deep features $P_5$ and $P_6$ extracted by the backbone. Meanwhile, ORSA enhances the shallow encoder features by making foreground visual representations more sensitive to textual semantics. During inference, QPSC, the distillation branch, and ORSA are all discarded.}
  \label{fig1}
\end{figure}

\section{Method}
\label{method}
In this section, we present the overall framework of VL-DINO, as illustrated in Figure \ref{fig1}, which consists of three core modules. QPSC is first introduced in Section~\ref{qpsc} to provide high-quality positive samples, followed by VSE in Section~\ref{vse}, which injects CLIP-derived visual semantics into backbone-extracted features to produce fused representations. Finally, ORSA in Section~\ref{orsa} further refines these representations by aligning object-region features with textual embeddings.

\subsection{Query-guided positive sample construction}
\label{qpsc}

As discussed in Section\ref{intro}, the vanilla DINO framework~\cite{dino17} is not well suited for the OVD setting. Inspired by the denoising mechanism~\cite{dino17} and the one-to-many training paradigm~\cite{group-detr23,hdetr24,co-detr25,deim26}, we propose QPSC, which constructs high-quality positive samples from reliable content queries to enhance vision-language alignment during training.

\begin{figure}[t]
  \centering
  % --- 左子图 (a) ---
  \begin{subfigure}{0.48\textwidth}
    \centering
    \includegraphics[width=\linewidth]{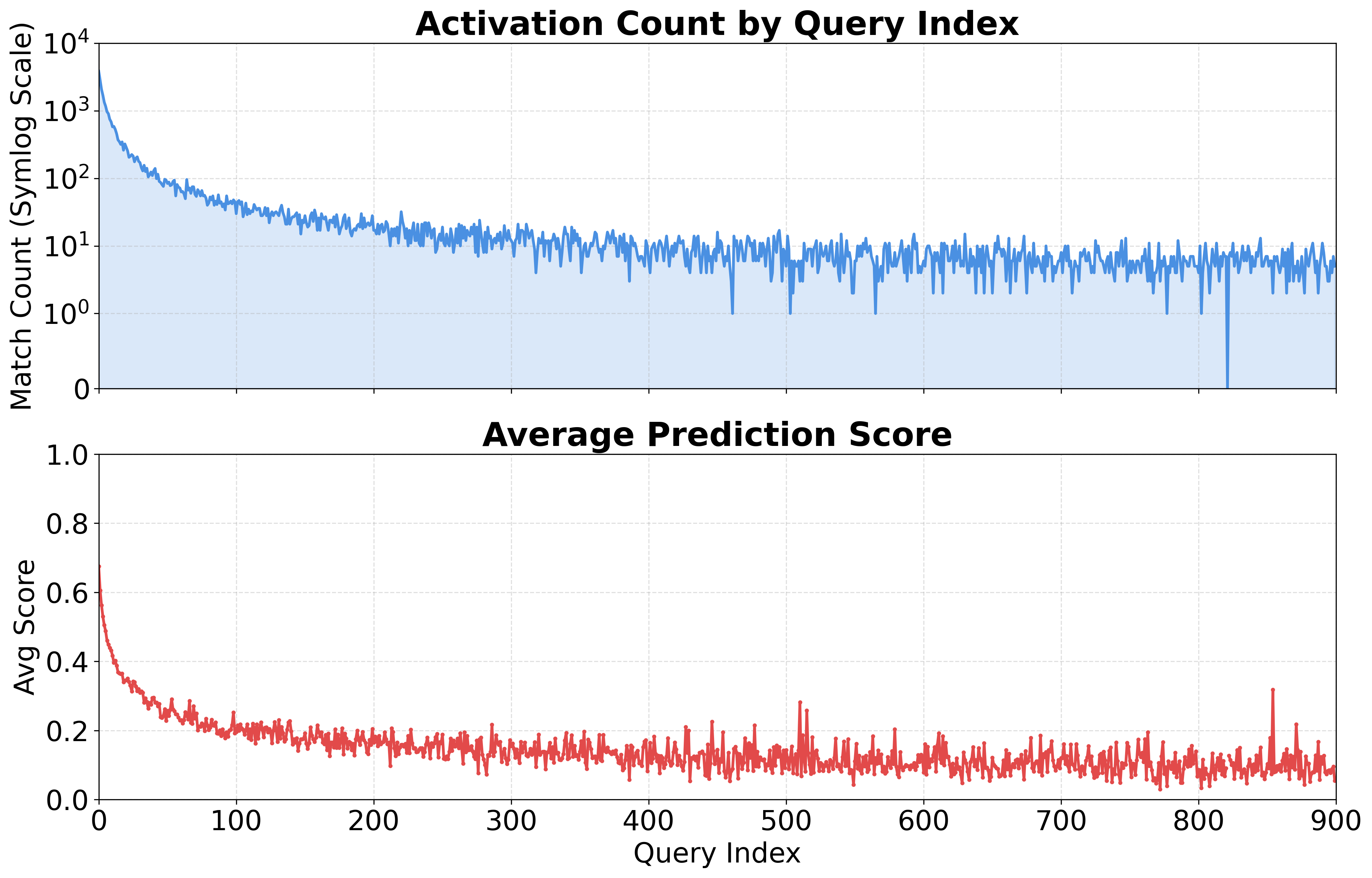}
    \caption{}
    \label{fig:query_coco} % 子图 a 的标签
  \end{subfigure}
  \hfill % 在两个子图之间插入弹性间距
  % --- 右子图 (b) ---
  \begin{subfigure}{0.48\textwidth}
    \centering
    \includegraphics[width=\linewidth]{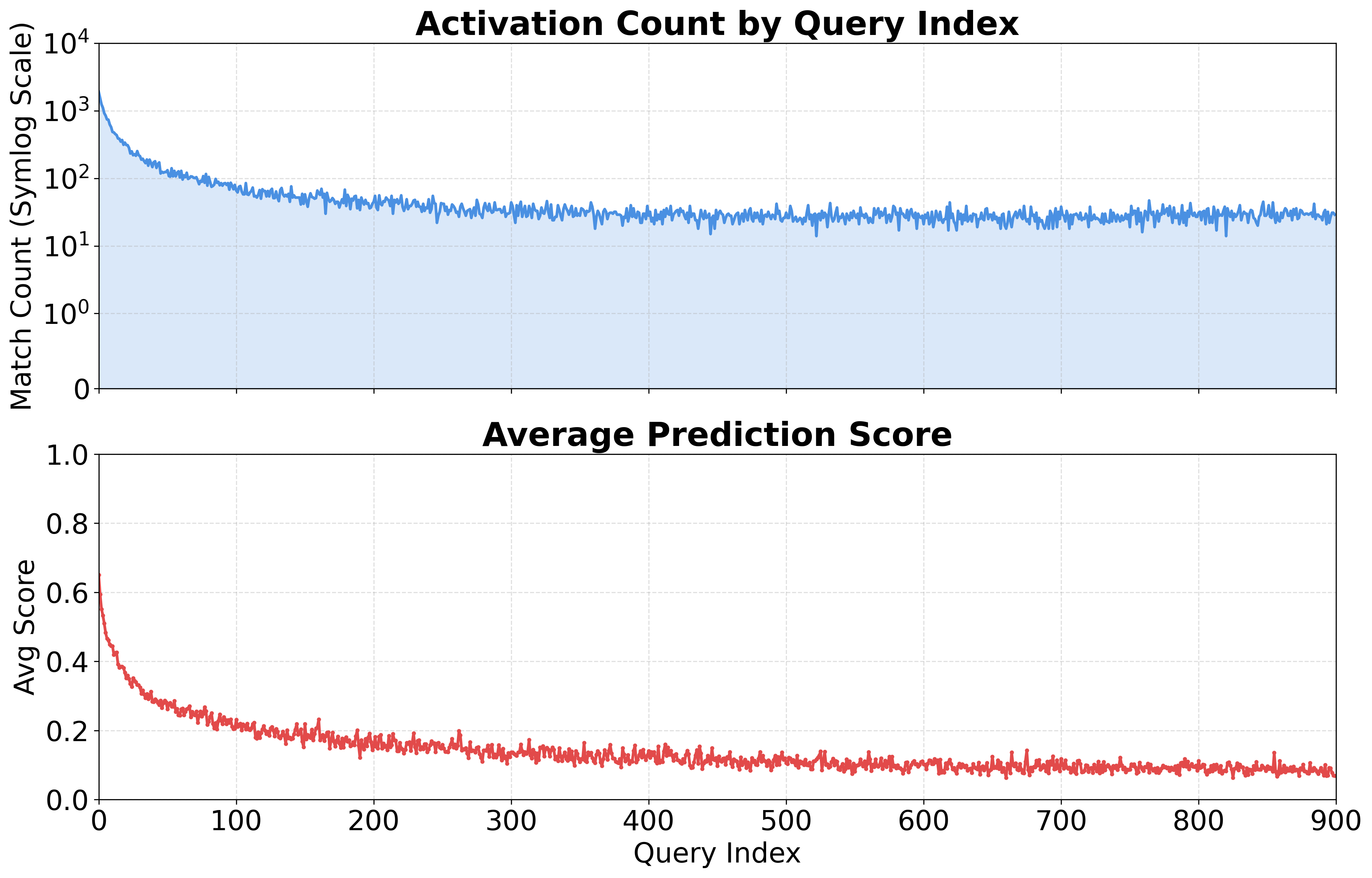}
    \caption{}
    \label{fig:query_lvis} % 子图 b 的标签
  \end{subfigure}

  % 全图的总标题
  \caption{Matching statistics of object queries. The model is trained on Objects365v1 and evaluated on (a) COCO \texttt{val2017} split and (b) LVIS \texttt{minival} split. In each subfigure, the top plot shows the total matches per query, while the bottom plot shows the average prediction score per query; both exhibit a clear long-tailed distribution.}
  \label{fig:overall_comparison} % 整张大图的标签

\end{figure}

To analyze the behavior of object queries, we conduct a preliminary experiment by training DINO model without the denoising mechanism on Objects365v1 dataset~\cite{objects365} until convergence. We then evaluate the model on COCO \texttt{val2017} split~\cite{COCO} and LVIS \texttt{minival} split~\cite{lvis}. Using the same Hungarian matching strategy as in training, we record, for each query, the total number of matches and its average prediction score. As shown in Figure \ref{fig:overall_comparison}, the matching statistics of object queries show a pronounced long-tailed distribution on both validation sets. Specifically, most matched predictions are concentrated within the first 300 queries, and these queries also yield higher average prediction scores. We hypothesize this behavior stems from the architectural bias of DINO: during training, predictions with higher foreground confidence tend to be implicitly associated with lower-indexed queries, allowing these queries to learn more reliable semantic representations. Based on this observation, the first 300 content queries are selected to construct a reliable query pool.

Following the box denoising in DINO~\cite{dino17}, QPSC applies bounded perturbations to each ground-truth box and generates an equal number of auxiliary boxes whose IoU with the corresponding ground truth is greater than 0.5. Each auxiliary box is paired with a content query embedding randomly sampled from the reliable query pool, forming an additional positive object query. These constructed positive object queries are fed into the decoder together with the original object queries and are supervised by both classification and box regression losses.
 
Notably, each of positive object queries is directly associated with its corresponding ground-truth object, thereby bypassing Hungarian matching during training, and are discarded during inference to avoid any extra computational burden. Furthermore, QPSC focuses exclusively on positive sample construction, while negative supervision remains handled by the original object queries. As a result, model receives richer positive supervision, leading to improved vision-language alignment.

\subsection{Visual semantic encoder}
\label{vse}

We argue that the multi-scale dense features extracted by the Swin Transformer backbone~\cite{swin20} are already highly expressive, and that its two deepest feature maps provide suitable carriers for visual semantic learning. Accordingly, we introduce VSE, composed of three self-attention blocks following the identical architecture of ViT~\cite{vit50} in CLIP~\cite{Radford1} and a multi-layer perceptron module. VSE operates on the $P_5$ and $P_6$ feature levels to enhance their global semantic capacity. Following CLIP, we prepend a learnable class embedding ($CLS$) to the input sequence. For each layer $l \in \{5,6\}$, we input the global class token $\mathrm{CLS}$ and the visual feature $P_l$ to obtain
\begin{equation}
\begin{gathered}
CLS_l,\ \bar{P}_l = \operatorname{SelfAttnBlcok}(CLS, P_l), \\
\tilde{P}_l = P_l + \operatorname{MLP}(\bar{P}_l)
\end{gathered}
\end{equation}
where $CLS_l$ is the output class embedding, $\bar{P}_l$ and $\tilde{P}_l$ are the visually enhanced feature and the fused feature, respectively. In particular, we treat $CLS_l$ as a proxy for distilling CLIP's visual semantic knowledge into $\bar{P}_l$. We first optimize a global semantic alignment loss $\mathcal{L}_{\mathrm{sem}}$ to learn semantic similarity between $CLS_l$ and the corresponding CLIP class embedding $CLS_{clip}$. In addition, we introduce an intra-batch relational loss $\mathcal{L}_{\mathrm{rel}}$ to capture the relative relationships among image samples within the same batch. The two losses are formulated as follows:
\begin{equation}
\begin{gathered}
\mathcal{L}_{sem}^l = 1 - \frac{1}{B} \sum_{i=1}^{B} \cos(\|CLS_{l}^{(i)}\|_2, \|CLS_{clip}^{(i)}\|_2), \\
\mathcal{L}_{rel}^l = \frac{1}{B} \sum_{i=1}^{B} KL \left( r_{clip}^{(i)} \parallel r_{l}^{(i)} \right),\\
r_c^{i,j} = \frac{\exp(\cos(\|CLS_c^{(i)}\|_2, \|CLS_c^{(j)}\|_2) / \tau_1)}{\sum_{k=1, k \neq i}^{B} \exp(\cos(\|CLS_c^{(i)}\|_2, \|CLS_c^{(k)}\|_2) / \tau_1)}, \quad j \neq i, c \in \{l, clip\}, \\ 
\end{gathered}
\label{eq:vse1}
\end{equation}
Here, $i$, $j$, and $k$ are sample indices within the batch, and $B$ is the batch size. $\tau_1$ is a temperature parameter. $KL(\cdot)$, $\cos(\cdot,\cdot)$, and $\|\cdot\|_2$ denote the Kullback--Leibler divergence, the cosine similarity, and the $L_2$ norm, respectively.

In addition to the aforementioned image-level losses, we further develop a dense spatial semantic loss $\mathcal{L}_{\mathrm{spa}}$ to model the spatial semantic relationships between patch tokens and the class embedding:

\begin{equation}
\begin{gathered}
\mathcal{L}_{spa}^l = \frac{1}{B} \sum_{i=1}^{B} KL \left( s_{clip} \parallel s_{l} \right), \\
s_c^{m} = \frac{\exp(\cos(\|CLS_c\|_2, \|\bar{P}_c^{m}\|_2) / \tau_2)}{\sum_{n=1}^{H_l \times W_l} \exp(\cos(\|CLS_c\|_2, \|\bar{P}_c^{n}\|_2) / \tau_2)}, \quad c \in \{l, clip\},\\
\end{gathered}
\label{eq:vse2}
\end{equation}

where $H_l \times W_l$ denotes the feature map of spatial size, and $m$ and $n$ index tokens. $\tau_2$ is a temperature parameter. In practice, we apply interpolation to $\bar{P}_{\mathrm{clip}}$ to match its spatial resolution with that of $\bar{P}_l$.

The overall distillation loss is defined as
\begin{equation}
\mathcal{L}_{\mathrm{dis}} = \frac{1}{2}(\lambda_{\mathrm{sem}} (\mathcal{L}_{\mathrm{sem}}^6+\mathcal{L}_{\mathrm{sem}}^5) + \lambda_{\mathrm{rel}} (\mathcal{L}_{\mathrm{rel}}^6+\mathcal{L}_{\mathrm{rel}}^5) + \lambda_{\mathrm{spa}} (\mathcal{L}_{\mathrm{spa}}^6+\mathcal{L}_{\mathrm{spa}}^5)),
\label{eq:dis}
\end{equation}
where $\lambda_{\mathrm{sem}}$, $\lambda_{\mathrm{rel}}$, and $\lambda_{\mathrm{spa}}$ are weighting coefficients that balance the contributions of the three loss terms. The features $\tilde{P}_5$ and $\tilde{P}_6$, enriched with CLIP visual semantic knowledge, serve as global image representations. Together with $P_3$ and $P_4$, they form a new set of multi-scale features, which are fed into the encoder to further refine dense visual representations.

\subsection{Object-region semantic alignment}
\label{orsa}

We design ORSA to better adapt the fused feature $\tilde{P}_l$ to detection while further incorporating textual semantic knowledge. Specifically, ORSA consists of a set of learnable queries and a foreground decoder, which follows the same architecture as the decoder but contains only three layers.

During training, the queries are guided by ground-truth box information to extract foreground object features from outputs of the first encoder layer. These features are then aligned with their corresponding textual features. During inference, the ORSA module is removed.

\section{Experiments}
\label{experiment}

In this section, we first introduce the implementation details, including the model configuration, training datasets, and evaluation metrics.We then present extensive experimental results and comprehensive analyses to demonstrate the effectiveness of the proposed method and to further understand the contribution of each component.

\subsection{Implementation details}
\label{implementation details}
\noindent\textbf{Training.} VL-DINO is built upon DINO~\cite{dino17} as the detection framework. The overall architecture consists of a Swin Transformer backbone~\cite{swin20}, six deformable encoder layers~\cite{deformable-detr28}, six deformable decoder layers~\cite{deformable-detr28}, and 900 object queries. For the language and the visual knowledge distillation branch, we use the pre-trained text and vision encoder from CLIP-B/16~\cite{Radford1}, whose parameters are frozen throughout training. Unless otherwise specified, we train VL-DINO on eight NVIDIA L40 GPUs using Objects365v1~\cite{objects365} and GoldG~\cite{mdetr22} consistent with previous methods. For simplicity, we refer to the two datasets as O and G, respectively.

\noindent\textbf{Evaluation.} To assess the open-vocabulary detection performance of the pre-trained VL-DINO, we perform zero-shot evaluation on the COCO~\citep{COCO} \texttt{val2017} split and the LVIS~\cite{lvis} \texttt{minival} split. These two benchmarks exhibit distinct class distributions and evaluation characteristics, and thus provide a comprehensive test bed for assessing the generalization ability of open-vocabulary detectors. For COCO, which contains 80 categories, we report the standard AP metric. For LVIS, which contains 1,203 categories, we report both standard AP and Fixed AP~\cite{fixed-ap42}, with the maximum number of predictions per image set to 300 and 900, respectively.

\begin{table*}[t]
  \caption{Zero-shot object detection performance comparison on LVIS. Cap4M denotes the dataset proposed in GLIP~\cite{glip7}, and G-20M denotes the Grounding-20M dataset~\cite{gdino1.5-44}.}
  \label{table1}
  \vspace{0.5\baselineskip}
  \centering
  \footnotesize
  \renewcommand{\arraystretch}{1}
  \setlength{\tabcolsep}{2pt}
  \renewcommand{\tabularxcolumn}[1]{m{#1}}
  \begin{tabularx}{\textwidth}{l c >{\centering\arraybackslash}X *{4}{r@{\,/\,}l}}
    \toprule
    Method & Backbone & Training Data
    & \multicolumn{2}{c}{$\mathrm{AP} / \mathrm{AP}^{\mathrm{Fixed}}$}
    & \multicolumn{2}{c}{$\mathrm{AP}_r / \mathrm{AP}^{\mathrm{Fixed}}_r$}
    & \multicolumn{2}{c}{$\mathrm{AP}_c / \mathrm{AP}^{\mathrm{Fixed}}_c$}
    & \multicolumn{2}{c}{$\mathrm{AP}_f / \mathrm{AP}^{\mathrm{Fixed}}_f$} \\
    \midrule
    GLIP-T~\cite{glip7} & Swin-T & OG & - & 24.9 & - & 17.7 & - & 19.5 & - & 31.0 \\
    GLIP-T~\cite{glip7} & Swin-T & OG,Cap4M & - & 26.0 & - & 20.8 & - & 21.4 & - & 31.0 \\
    GLIPv2-T~\cite{glipv2-43} & Swin-T & OG,Cap4M & - & 29.0 & - & - & - & - & - & - \\
    GDINO-T~\cite{gdino6} & Swin-T & OG & - & 25.6 & - & 14.4 & - & 19.6 & - & 32.2 \\
    GDINO-T~\cite{gdino6} & Swin-T & OG,Cap4M & - & 27.4 & - & 18.1 & - & 23.3 & - & 32.7 \\
    G1.5-Edge~\cite{gdino1.5-44} & EfficientViT-L1 & G-20M & - & 33.5 & - & 28.0 & - & 34.3 & - & 33.9 \\
    T-Rex2-T~\cite{T-rex2-5} & Swin-T & OG & - & 34.9 & - & 32.7 & - & 32.9 & - & 37.1 \\
    YWorldv2-S~\cite{yolo-world3} & YOLOv8-S & OG & - & 24.4 & - & 17.1 & - & 22.5 & - & 27.3 \\
    YWorldv2-M~\cite{yolo-world3} & YOLOv8-M & OG & - & 32.4 & - & 28.4 & - & 29.6 & - & 35.5 \\
    YWorldv2-L~\cite{yolo-world3} & YOLOv8-L & OG & - & 35.5 & - & 25.6 & - & 34.6 & - & 38.1 \\
    YOLOEv8-S~\cite{yoloe4} & YOLOv8-S & OG & 25.7 & 27.9 & 19.0 & 22.3 & 25.9 & 27.8 & 26.7 & 29.0 \\
    YOLOEv8-M~\cite{yoloe4} & YOLOv8-M & OG & 29.9 & 32.6 & 23.6 & 26.9 & 29.2 & 31.9 & 31.7 & 34.4 \\
    YOLOEv8-L~\cite{yoloe4} & YOLOv8-L & OG & 33.3 & 35.9 & 30.8 & 33.2 & 32.2 & 34.8 & 34.6 & 37.3 \\
    YOLOEv11-S~\cite{yoloe4} & YOLOv8-S & OG & 25.2 & 27.5 & 19.3 & 21.4 & 24.4 & 26.8 & 26.7 & 29.3 \\
    YOLOEv11-M~\cite{yoloe4} & YOLOv8-M & OG & 30.5 & 33.0 & 22.4 & 26.9 & 30.4 & 32.5 & 32.1 & 34.5 \\
    YOLOEv11-L~\cite{yoloe4} & YOLOv8-L & OG & 32.4 & 35.2 & 25.6 & 29.1 & 31.9 & 35.0 & 34.1 & 36.5 \\
    OV-DEIM-S~\cite{ov-deim45} & ViT-T & OG & 27.7 & 29.6 & 23.6 & 25.2 & 28.1 & 30.2 & 28.0 & 30.0 \\
    OV-DEIM-M~\cite{ov-deim45} & ViT-T+ & OG & 30.6 & 32.6 & 25.3 & 26.9 & 30.2 & 31.5 & 31.9 & 34.1 \\
    OV-DEIM-L~\cite{ov-deim45} & ViT-S & OG & 33.7 & 35.9 & 34.3 & 36.8 & 33.4 & 35.5 & 34.0 & 36.0 \\
    \midrule
    VL-DINO-T (ours) & Swin-T & OG & 36.3 & 38.7 & 37.8 & 40.6 & 35.6 & 37.9 & 36.7 & 39.0 \\
    VL-DINO-L (ours) & Swin-L & OG & 38.1 & 40.2 & 39.2 & 41.8 & 38.1 & 40.9 & 37.7 & 40.1 \\
    \bottomrule
  \end{tabularx}
\end{table*}

\begin{figure}[t]
    \centering
    
    % 第一张子图
    \begin{subfigure}{0.32\textwidth}
        \centering
        % 请将 image1.png 替换为你上传的实际文件名
        \includegraphics[width=\textwidth]{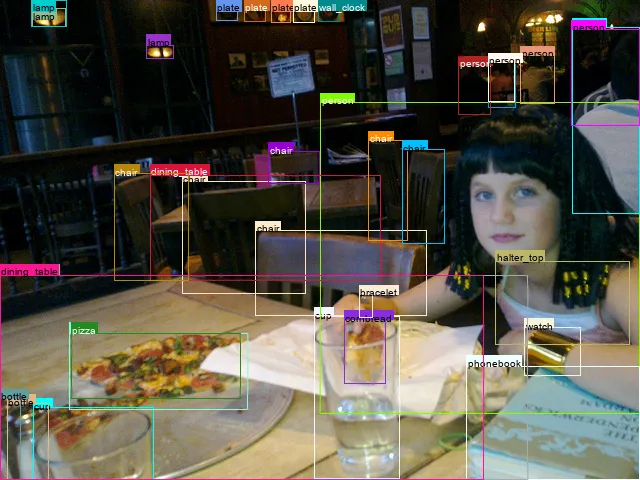}
        \caption{}
        \label{fig:sub1}
    \end{subfigure}
    \hfill % 添加弹性间距，让图片自动均匀散开
    % 第二张子图
    \begin{subfigure}{0.32\textwidth}
        \centering
        \includegraphics[width=\textwidth]{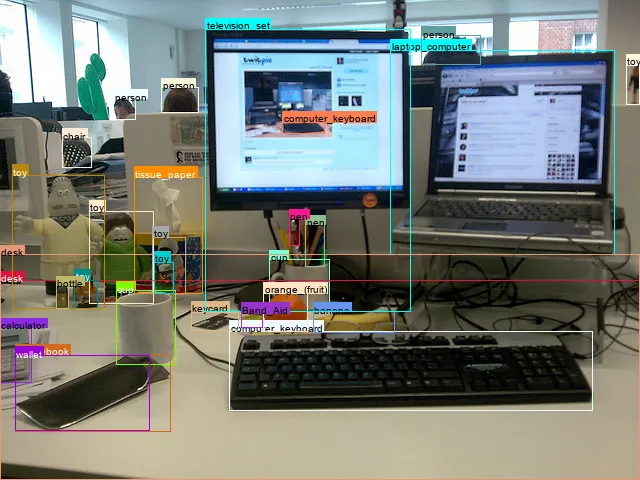}
        \caption{}
        \label{fig:sub2}
    \end{subfigure}
    \hfill
    % 第三张子图
    \begin{subfigure}{0.32\textwidth}
        \centering
        \includegraphics[width=\textwidth]{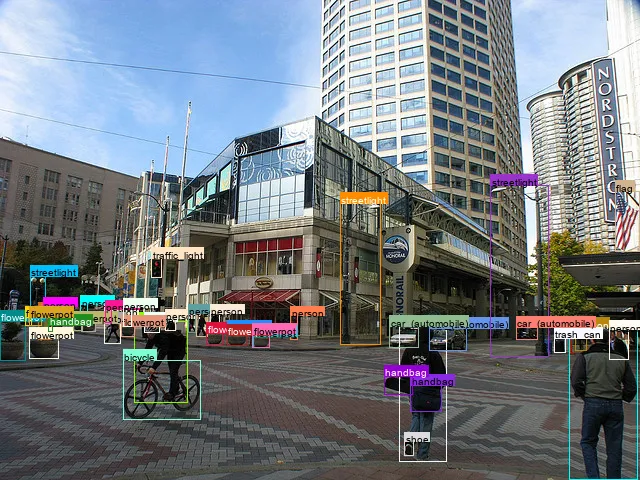}
        \caption{}
        \label{fig:sub3}
    \end{subfigure}

    \caption{Visualizations of zero-shot inference on LVIS. The pretrained VL-DINO-T model is evaluated with all 1,203 category text embeddings as input.}
    \label{fig:vis}
    \vspace{-0.8\baselineskip}
\end{figure}

\subsection{Zero-shot results}
\noindent\textbf{LVIS.} As shown in Table \ref{table1}, VL-DINO achieves the best performance on LVIS in the zero-shot setting. VL-DINO-T obtains 36.3 AP and 38.7 Fixed AP, while VL-DINO-L further reaches 38.1 AP and 40.2 Fixed AP, surpassing all advanced models under this setting. Compared with GLIP-T and GDINO-T, VL-DINO-T achieves substantially better performance without resorting to their computationally intensive layer-wise cross-modal fusion modules or large-scale training corpora. At inference time, our method retains only extra VSE, yet still surpasses GLIP-T and GDINO-T by 12.7 and 11.3 Fixed AP, respectively. VL-DINO-T also outperforms recent YOLO-style detectors(e.g., YWorld and YOLOE) and DINOv3-style~\cite{dinov3-46} detector(e.g., OV-DEIM), exceeding YOLOEv8-S by 10.8 Fixed AP and OV-DEIM-S by 9.1 Fixed AP, while consistently outperforming all reported variants. Moreover, on the challenging rare categories of LVIS, VL-DINO-L achieves the highest 41.8 Fixed AP, highlighting its strong effectiveness on long-tailed open-vocabulary detection. We present qualitative visualization results in Figure ~\ref{fig:vis}, demonstrating that VL-DINO-T can accurately localize multiple relevant objects across diverse scenes, while maintaining robust performance under varying object scales and complex visual contexts.

\noindent\textbf{COCO.} COCO mainly contains object categories with a relatively balanced class distribution, and thus serves as an important benchmark for evaluating zero-shot detection performance on frequent classes. As shown in Table ~\ref{table2}, under the same OG training data, VL-DINO achieves strong performance on COCO in the zero-shot setting, with VL-DINO-T and VL-DINO-L reaching 48.5 and 50.0 AP, respectively. In particular, VL-DINO-T outperforms GLIP-T, GDINO-T, and T-Rex2-T under the same Swin-T backbone and OG training setup. Compared with recent YWorldv1 and OV-DEIM detectors, VL-DINO-T also shows clear advantages, surpassing YWorldv1-L by 4.1 AP and OV-DEIM-L by 2.6 AP. These results demonstrate that VL-DINO not only maintains strong open-vocabulary generalization, but also exhibits superior detection capability on frequent categories.

\begin{table}[t]
  \centering
  
  % =================== 左侧：主性能表格 ===================
  % 【修改 1】：宽度调为 0.46，给变宽的右侧让路
  \begin{minipage}[t]{0.46\linewidth}
    \vspace{0pt} 
    \captionof{table}{Zero-shot object detection performance comparison on COCO.}
    \label{table2}
    \centering
    \scriptsize 
    
    \renewcommand{\arraystretch}{1.185} 
    \setlength{\tabcolsep}{1.5pt} 
    \renewcommand{\tabularxcolumn}[1]{m{#1}}
    
    \begin{tabularx}{\linewidth}{>{\raggedright\arraybackslash}X c >{\centering\arraybackslash}X c }
      \toprule
      Model & Backbone & Data & AP \\
      \midrule
      GLIP-T~\cite{glip7}  & Swin-T & OG & 46.7  \\
      GDINO-T~\cite{gdino6} & Swin-T & OG & 48.1  \\
      GDINO-L~\cite{gdino6} & Swin-L & OG,OI & 52.5  \\
      T-Rex2-T~\cite{T-rex2-5} & Swin-T & OG & 46.4 \\
      YWorldv1-S~\cite{yolo-world3} & YOLOv8-S & OG & 37.6 \\
      YWorldv1-M~\cite{yolo-world3} & YOLOv8-M & OG & 42.8 \\
      YWorldv1-L~\cite{yolo-world3} & YOLOv8-L & OG & 44.4 \\
      OV-DEIM-S~\cite{ov-deim45} & ViT-T & OG & 40.8 \\
      OV-DEIM-M~\cite{ov-deim45} & ViT-T+ & OG & 43.3 \\
      OV-DEIM-L~\cite{ov-deim45} & ViT-S & OG & 45.9 \\
      \midrule
      VL-DINO-T (ours) & Swin-T & OG & 48.5 \\
      VL-DINO-L (ours) & Swin-L & OG & 50.0 \\
      \bottomrule
    \end{tabularx}
  \end{minipage}
  \hfill 
  %
  % =================== 右侧：上下堆叠的两个表格 ===================
  % 【修改 1】：宽度调为 0.51。0.46 + 0.51 = 0.97，完美并排放下
  \begin{minipage}[t]{0.51\linewidth}
    \vspace{0pt} 
    
    % --- 右侧上部分：消融实验表格 (Table 3) ---
    \captionof{table}{Ablation study of QPSC, VSE, and ORSA, where modules are progressively introduced.}
    \label{table3}
    \centering
    \scriptsize 
    \renewcommand{\arraystretch}{1.12} 
    \setlength{\tabcolsep}{1.5pt} % 7 列比较紧凑，稍微收紧内边距
    
    % 【修改 2】：匹配 7 列数据，将 *{5} 改为 *{7}
    \begin{tabularx}{\linewidth}{*{7}{>{\centering\arraybackslash}X}}
      \toprule
       QPC & VSE & ORSA & AP & $\mathrm{AP}_{\mathrm{r}}$ & $\mathrm{AP}_{\mathrm{c}}$ & $\mathrm{AP}_{\mathrm{f}}$ \\
      \midrule
      \ding{55} & \ding{55} & \ding{55} & 31.8 & 26.3 & 31.2 & 33.4 \\
      \ding{51} & \ding{55} & \ding{55} & 33.5 & 30.1 & 32.4 & 35.0 \\
      \ding{51} & \ding{51} & \ding{55} & 35.6 & 35.1 & 35.8 & 35.6 \\
      \ding{51} & \ding{51} & \ding{51} & 36.3 & 37.8 & 35.6 & 36.7 \\
      \bottomrule
    \end{tabularx}
    
    % 合并冗余间距，统一调整为 0.35cm
    \vspace{0.35cm} 
    
   % --- 右侧下部分：新增表格 (Table 4) ---
    \captionof{table}{Effect of QPSC on LVIS and COCO.}
    \vspace{-0.1cm}
    \label{table4}
    \centering
    \scriptsize 
    \renewcommand{\arraystretch}{1.15} 
    \setlength{\tabcolsep}{2pt}
    
    \begin{tabular*}{0.9\linewidth}{@{\extracolsep{\fill}} c c c c @{}}
      \toprule
      QPSC & Training Data & $\mathrm{AP}_{\mathrm{COCO}}$ & $\mathrm{AP}_{\mathrm{LVIS}}$ \\
      \midrule
      \ding{55} & O & 42.9 & 25.3 \\
      \ding{51} & O & 44.9 & 25.7 \\
      \ding{55} & OG & 44.5 & 31.8 \\
      \ding{51} & OG & 45.4 & 33.5 \\
      \bottomrule
    \end{tabular*}
    
  \end{minipage}

\end{table}

\subsection{Ablation study}

To better understand the contribution of each design in VL-DINO, we perform a series of ablation studies on QPSC, VSE, and ORSA. Unless stated otherwise, all ablation experiments are conducted using the VL-DINO-T trained on OG dataset and evaluated on LVIS in the zero-shot setting.

\noindent\textbf{Overall component analysis.} We first conduct a progressive component analysis on LVIS by incrementally introducing each proposed module into the baseline, so as to evaluate its individual contribution to the final performance. As shown in Table \ref{table3}, incorporating QPSC and VSE yields gains of 1.7 AP and 2.1 AP, respectively. In particular, compared with the baseline, the AP$_r$ score improves by 8.8 AP, suggesting that the richer vision-language supervision brought by QPSC and the visual semantic knowledge learned through VSE are especially effective for rare-category detection. It is also worth noting that, although ORSA improves the overall AP by only 0.7, it further boosts AP$_r$ by 2.7 AP. This result indicates that the foreground object-region alignment in ORSA effectively encourages the model to learn more discriminative representations for rare categories.

\noindent\textbf{Ablation study of QPSC under different training data settings.} We analyze the effect of QPSC under different training data settings. As shown in Table \ref{table4}, introducing QPSC consistently improves the zero-shot detection performance on both COCO and LVIS, regardless of whether the model is trained on O or OG. The gains are particularly noticeable on LVIS under the OG setting, where QPSC improves the AP from 31.8 to 33.5. These results verify that QPSC provides effective additional supervision and improves the model's ability to learn transferable vision-language semantics.

\noindent\textbf{Ablation study of VSE under different distillation losses.} As shown in Table ~\ref{table5}, introducing VSE alone, without distillation supervision, provides limited benefit. Specifically, applying VSE only to $\mathcal{P}_6$ achieves 33.7 AP, whereas extending it to both $\mathcal{P}_6$ and $\mathcal{P}_5$ slightly decreases the performance to 33.3 AP, with consistent drops in AP$_r$, AP$_c$, and AP$_f$. This suggests that VSE itself is insufficient to improve features, and may even introduce interference when no explicit semantic supervision is imposed. Once distillation losses are applied, however, the effectiveness of VSE becomes clear. Using $\mathcal{L}_{sem}$ and $\mathcal{L}_{rel}$ in Eq \ref{eq:vse1} already improves the performance, especially when both $\mathcal{P}_6$ and $\mathcal{P}_5$ are distilled jointly, where the AP rises to 35.3 and AP$_r$ reaches 35.0, indicating that semantic and relational supervision effectively facilitate the transfer of CLIP visual knowledge. Further adding the spatial distillation loss $\mathcal{L}_{spa}$ in Eq \ref{eq:vse2} leads to the best overall result of 35.6 AP, while also achieving the highest AP$_r$ of 35.1 and AP$_c$ of 35.8. Overall, these results show that the gains of VSE mainly come from the proposed distillation objectives rather than vse structure itself, and that semantic, relational, and spatial supervision are complementary for learning richer visual semantic knowledge, especially on rare categories.

\noindent\textbf{Ablation study of the number of self-attention blocks in VSE.} Table ~\ref{table6} reports the effect of varying the number of self-attention blocks in VSE. As the number increases from 1 to 3, the overall AP improves steadily from 34.7 to 35.6. In particular, AP$_r$ rises from 33.8 to 35.1, suggesting that a larger VSE is especially helpful for transferring visual semantics to rare-category detection. Although the 2-block setting achieves the highest AP$_f$ of 36.1, the 3-block setting delivers the best overall AP together with the best AP$_r$ and AP$_c$, indicating a better balance across different category groups. These results suggest that a shallow VSE is insufficient to fully model CLIP's high-level visual semantics, while the 3-block design provides stronger semantic encoding capacity and is therefore adopted in our final model.

\noindent\textbf{Ablation study of alignment loss in ORSA.} Starting from the best VSE setting, which achieves 35.6 AP, 35.1 AP$_r$, 35.8 AP$_c$, and 35.6 AP$_f$, introducing ORSA with the initial focal loss~\cite{focal-loss31} already brings further improvement to 36.0 AP and 35.6 AP$_r$, as shown in Table ~\ref{table7}. This choice is reasonable because ORSA extracts holistic object-level foreground features, for which classification-style supervision can serve as a simple alignment objective. We then replace focal loss with contrastive alignment loss~\cite{infonce-loss-49}. Although the overall AP changes only slightly, image-to-text loss improves AP$_r$ to 36.7, and the bidirectional loss further boosts AP$_r$ to 37.8 while achieving the best overall AP of 36.3. In comparison, the changes on AP$_c$ and AP$_f$ are relatively small, indicating that the benefit of contrastive supervision is much more pronounced on rare categories than on common or frequent ones. This suggests that ORSA mainly helps the model learn more discriminative semantic representations for challenging long-tail classes. Based on these results, we adopt the bidirectional contrastive loss as the default alignment loss in ORSA.

\begin{table*}[t]
  \centering
  
  % 【核心且唯一的一步】：直接限制表头的宽度，让它和下面自然舒展的表格对齐
  \captionsetup{width=0.65\textwidth}
  \caption{Ablation study of VSE under different distillation loss settings and feature maps.}
  \label{table5}
  \vspace{0.5\baselineskip}
  
  \footnotesize
  \renewcommand{\arraystretch}{1.1}
  \setlength{\tabcolsep}{10pt} 
  
  \begin{tabular}{l c c c c c}
    \toprule
    Distillation Loss & Feature Map & AP & $\mathrm{AP}_r$ & $\mathrm{AP}_c$ & $\mathrm{AP}_f$ \\
    \midrule
    -- & $\mathcal{P}_{6}$ & 33.7 & 30.0 & 32.5 & 35.2 \\
    -- & $\mathcal{P}_{6}, \mathcal{P}_{5}$ & 33.3 & 29.4 & 32.0 & 34.8 \\
    $\mathcal{L}_{sem}, \mathcal{L}_{rel}$ & $\mathcal{P}_{6}$ & 34.3 & 30.1 & 33.4 & 35.7 \\
    $\mathcal{L}_{sem}, \mathcal{L}_{rel}$ & $\mathcal{P}_{6}, \mathcal{P}_{5}$ & 35.3 & 35.0 & 34.6 & 35.8 \\
    $\mathcal{L}_{sem}, \mathcal{L}_{rel}, \mathcal{L}_{spa}$ & $\mathcal{P}_{6}$ & 34.7 & 34.1 & 33.8 & 35.8 \\
    $\mathcal{L}_{sem}, \mathcal{L}_{rel}, \mathcal{L}_{spa}$ & $\mathcal{P}_{6}, \mathcal{P}_{5}$ & 35.6 & 35.1 & 35.8 & 35.6 \\
    \bottomrule
  \end{tabular}
  \vspace{-0.8\baselineskip}
\end{table*}

\begin{table*}[t]
  \centering
  
  % --- 左侧领地：严格占据页面 48% 的宽度 ---
  \begin{minipage}[t]{0.48\linewidth}
    \centering
    \caption{Ablation study of the number of self-attention blocks in VSE.}
    \label{table6}
    \vspace{0.5\baselineskip}
    \footnotesize
    \renewcommand{\arraystretch}{1.2}
    
    % 将第一个 l 改成了 c，现在四列全部居中对齐
    \begin{tabular*}{\linewidth}{@{\extracolsep{\fill}} c c c c c @{}}
      \toprule
      Block Num & $\mathrm{AP}$ & $\mathrm{AP}_r$ & $\mathrm{AP}_c$ & $\mathrm{AP}_f$ \\
      \midrule
      1 & 34.7 & 33.8 & 33.9 & 35.5 \\
      2 & 35.3 & 34.8 & 34.4 & 36.1 \\
      3 & 35.6 & 35.1 & 35.8 & 35.6 \\
      \bottomrule
    \end{tabular*}
  \end{minipage}
  \hfill
  % --- 右侧领地：严格占据页面 48% 的宽度 ---
  \begin{minipage}[t]{0.48\linewidth}
    \centering
    \caption{Ablation study of ORSA with different alignment losses.}
    \label{table7}
    \vspace{0.5\baselineskip}
    \footnotesize
    \renewcommand{\arraystretch}{1.2}
    
    % 同理，四列全部设为 c
    \begin{tabular*}{\linewidth}{@{\extracolsep{\fill}} c c c c c @{}}
      \toprule
      Alignment Loss & $\mathrm{AP}$ & $\mathrm{AP}_r$ & $\mathrm{AP}_c$ & $\mathrm{AP}_f$ \\
      \midrule
      $\mathcal{L}_{\mathrm{focal}}$ & 36.0 & 35.6 & 35.7 & 36.4 \\
      $\mathcal{L}_{\mathrm{i2t}}$ & 36.1 & 36.7 & 35.6 & 36.5 \\
      $\mathcal{L}_{\mathrm{i2t}} + \mathcal{L}_{\mathrm{t2i}}$ & 36.3 & 37.8 & 35.6 & 36.7 \\
      \bottomrule
    \end{tabular*}
  \end{minipage}

\end{table*}

\begin{table}[t]
  \centering
  
  % --- 左侧领地：严格占据页面 48% 的宽度 ---
  \begin{minipage}[t]{0.48\linewidth}
    \centering
    \caption{Ablation of context-aware box scaling in ORSA.}
    \label{table8}
    \footnotesize
    \renewcommand{\arraystretch}{1.3}
    
    % 将第一个 l 改成了 c，现在四列全部居中对齐
    \begin{tabular*}{\linewidth}{@{\extracolsep{\fill}} c c c c c @{}}
      \toprule
      Scale Factor & $\mathrm{AP}$ & $\mathrm{AP}_r$ & $\mathrm{AP}_c$ & $\mathrm{AP}_f$ \\
      \midrule
      $1.0\times$ & 36.3 & 37.8 & 35.6 & 36.7 \\
      $1.1\times$ & 36.1 & 36.6 & 35.9 & 36.4 \\
      $1.2\times$ & 35.8 & 36.1 & 34.9 & 36.5 \\
      \bottomrule
    \end{tabular*}
    \vspace{-0.8\baselineskip}
  \end{minipage}
  \hfill
  % --- 右侧领地：严格占据页面 48% 的宽度 ---
  \begin{minipage}[t]{0.48\linewidth}
    \centering
    \caption{Ablation study of the number of foreground decoder layers in ORSA.}
    \label{table9}
    \footnotesize
    \renewcommand{\arraystretch}{1.3}
    
    % 同理，四列全部设为 c
    \begin{tabular*}{\linewidth}{@{\extracolsep{\fill}} c c c c c @{}}
      \toprule
      Layer Num & $\mathrm{AP}$ & $\mathrm{AP}_r$ & $\mathrm{AP}_c$ & $\mathrm{AP}_f$ \\
      \midrule
      1 & 36.1 & 35.8 & 35.9 & 36.2 \\
      3 & 36.3 & 37.8 & 35.6 & 36.7 \\
      6 & 36.4 & 38.2 & 35.7 & 36.8 \\
      \bottomrule
    \end{tabular*}
    \vspace{-0.8\baselineskip}
  \end{minipage}

\end{table}

\noindent\textbf{Ablation study of box scaling in ORSA.} ORSA extracts object-region features using ground-truth box annotations. To examine whether introducing a moderate amount of surrounding context can improve vision-language alignment, we enlarge the width and height of the ground-truth boxes by different scale factors. As shown in Table ~\ref{table8}, the default setting without enlargement achieves the best overall performance, reaching 36.3 AP and 37.8 AP$_r$. Enlarging the box to $1.1\times$ or $1.2\times$ leads to consistent degradation in overall AP, with particularly clear drops on rare categories. Although the $1.1\times$ setting slightly improves AP$_c$, its overall performance remains inferior. We conjecture that enlarging the box introduces additional background regions, causing the deformable encoder layer to encode background noise into the object-region features, which weakens visual feature representation. Therefore, we adopt the original ground-truth box without enlargement.

\noindent\textbf{Ablation study of the number of foreground decoder layers in ORSA.} Table ~\ref{table9} reports the effect of varying the layer number in ORSA. Compared with the baseline without ORSA, introducing only 1 layer already improves the overall AP from 35.6 to 36.1, demonstrating that foreground object-region alignment is effective for enhancing detection. Increasing the layer number to 3 further boosts the performance to 36.3 AP, with a particularly notable gain on rare categories, where AP$_r$ reaches 37.8. This indicates that a deeper ORSA can provide stronger semantic refinement for challenging categories. Although extending ORSA to 6 layers yields the best numerical results, improving AP to 36.4 and AP$_r$ to 38.2, the gain over the 3-layer setting is relatively small. Considering that a deeper ORSA also introduces more parameters and computation, we adopt 3 layers as the default setting, which offers a better trade-off between performance and complexity.

\subsection{Transfer learning on COCO}
Following YOLOE~\cite{yoloe4}, we evaluate the transferability of pretrained models to COCO under both linear probing and full tuning settings, where linear probing only fine-tunes the classification layers while keeping the rest of the model frozen, and full tuning updates all model parameters. As shown in Table ~\ref{table10}, VL-DINO consistently achieves the best performance in both regimes, demonstrating strong transferability of the pretrained representations. Under linear probing, VL-DINO-T and VL-DINO-L reach 51.5 AP and 53.1 AP, respectively, outperforming all compared baselines by clear margins. Under full tuning, VL-DINO further improves to 55.9 AP and 57.5 AP. In addition to the gains in AP, VL-DINO also achieves the best AP$_{50}$ and AP$_{75}$, indicating that the transferred representations benefit both classification and localization. Overall, these results indicate that VL-DINO learns highly transferable visual-semantic features that can be better adapted to downstream detection tasks.

\begin{table}[t]
  \centering
  
  % 包裹在 threeparttable 中，表头会自动探测并对齐下方 tabular 的宽度
  \begin{threeparttable}
  \caption{Transfer learning results on COCO under both linear probing and full tuning settings.}
  \label{table10}
    \vspace{0.5\baselineskip}
  \footnotesize
  \renewcommand{\arraystretch}{1.1}
  \setlength{\tabcolsep}{10pt} 
  
  \begin{tabular}{l c c c c}
    \toprule
    % 修正：原代码有两个 AP_50，此处调整为标准 COCO 评价指标
    Model  & Epochs & $\mathrm{AP}$ & $\mathrm{AP}_{50}$ & $\mathrm{AP}_{75}$ \\
    \midrule
    
    % 修正：表格共 5 列，因此 \multicolumn 必须是 5 才能完美居中
    \multicolumn{5}{c}{\emph{Linear Probing}} \\
    YOLOEv11-S~\cite{yoloe4} & 10 & 37.0 & 52.9 & 40.4 \\
    YOLOEv11-M~\cite{yoloe4} & 10 & 43.1 & 60.6 & 47.4 \\
    YOLOEv11-L~\cite{yoloe4} & 10 & 45.1 & 62.8 & 49.5 \\
    VL-DINO-T (ours) & 10 & 51.5 & 67.6 & 56.9 \\
    VL-DINO-L (ours) & 10 & 53.1 & 69.8 & 58.8 \\

    % 修正：将 \hline 换为 \midrule，保持上下线型粗细风格一致
    \midrule
    
    \multicolumn{5}{c}{\emph{Full Tuning}} \\
    YOLOEv11-S~\cite{yoloe4} & 160 & 46.2 & 62.9 & 50.0 \\
    YOLOEv11-M~\cite{yoloe4} & 80 & 51.3 & 68.3 & 56.0 \\
    YOLOEv11-L~\cite{yoloe4} & 80 & 52.6 & 69.7 & 57.5 \\
    VL-DINO-T (ours) & 15 & 55.9 & 72.1 & 61.7 \\
    VL-DINO-L (ours) & 15 & 57.5 & 74.4 & 63.8 \\
    \bottomrule
  \end{tabular}
  \end{threeparttable}
\end{table}

\section{Conclusion and limitation}
\label{conclusion}
We present VL-DINO, an open-vocabulary detector that enhances DINO by jointly leveraging CLIP's textual and visual semantic knowledge. We introduce QPSC for richer textual supervision, VSE to distill CLIP visual features, and ORSA to align object-region representations with text. These designs effectively avoids over-reliance on a single modality and achieves stronger semantic transfer in open-vocabulary detection. VL-DINO offers an effective and scalable direction for integrating multimodal semantic priors into DETR-style open-vocabulary detectors. One limitation of our work is that the current framework does not incorporate more advanced visual foundation models, such as DINOv3~\cite{dinov3-46}, to learn richer visual semantic knowledge. Since stronger visual foundation models may provide more discriminative and transferable representations, exploring their integration into our framework is left for future work.

{
\small
\bibliographystyle{plainnat}
\bibliography{references}

\begin{thebibliography}{46}
\providecommand{\natexlab}[1]{#1}
\providecommand{\url}[1]{\texttt{#1}}
\expandafter\ifx\csname urlstyle\endcsname\relax
  \providecommand{\doi}[1]{doi: #1}\else
  \providecommand{\doi}{doi: \begingroup \urlstyle{rm}\Url}\fi

\bibitem[Carion et~al.(2020)Carion, Massa, Synnaeve, Usunier, Kirillov, and Zagoruyko]{detr27}
Nicolas Carion, Francisco Massa, Gabriel Synnaeve, Nicolas Usunier, Alexander Kirillov, and Sergey Zagoruyko.
\newblock End-to-end object detection with transformers.
\newblock In \emph{European conference on computer vision}, pages 213--229. Springer, 2020.

\bibitem[Chen et~al.(2023)Chen, Chen, Wang, Zhang, Yao, Feng, Han, Ding, Zeng, and Wang]{group-detr23}
Qiang Chen, Xiaokang Chen, Jian Wang, Shan Zhang, Kun Yao, Haocheng Feng, Junyu Han, Errui Ding, Gang Zeng, and Jingdong Wang.
\newblock Group detr: Fast detr training with group-wise one-to-many assignment.
\newblock In \emph{Proceedings of the IEEE/CVF international conference on computer vision}, pages 6633--6642, 2023.

\bibitem[Cheng et~al.(2024)Cheng, Song, Ge, Liu, Wang, and Shan]{yolo-world3}
Tianheng Cheng, Lin Song, Yixiao Ge, Wenyu Liu, Xinggang Wang, and Ying Shan.
\newblock Yolo-world: Real-time open-vocabulary object detection.
\newblock In \emph{Proceedings of the IEEE/CVF conference on computer vision and pattern recognition}, pages 16901--16911, 2024.

\bibitem[Dave et~al.(2021)Dave, Doll{\'a}r, Ramanan, Kirillov, and Girshick]{fixed-ap42}
Achal Dave, Piotr Doll{\'a}r, Deva Ramanan, Alexander Kirillov, and Ross Girshick.
\newblock Evaluating large-vocabulary object detectors: The devil is in the details.
\newblock \emph{arXiv preprint arXiv:2102.01066}, 2021.

\bibitem[Dosovitskiy et~al.(2020)Dosovitskiy, Beyer, Kolesnikov, Weissenborn, Zhai, Unterthiner, Dehghani, Minderer, Heigold, Gelly, et~al.]{vit50}
Alexey Dosovitskiy, Lucas Beyer, Alexander Kolesnikov, Dirk Weissenborn, Xiaohua Zhai, Thomas Unterthiner, Mostafa Dehghani, Matthias Minderer, Georg Heigold, Sylvain Gelly, et~al.
\newblock An image is worth 16x16 words: Transformers for image recognition at scale.
\newblock \emph{arXiv preprint arXiv:2010.11929}, 2020.

\bibitem[Du et~al.(2022)Du, Wei, Zhang, Shi, Gao, and Li]{detpro10}
Yu~Du, Fangyun Wei, Zihe Zhang, Miaojing Shi, Yue Gao, and Guoqi Li.
\newblock Learning to prompt for open-vocabulary object detection with vision-language model.
\newblock In \emph{Proceedings of the IEEE/CVF conference on computer vision and pattern recognition}, pages 14084--14093, 2022.

\bibitem[Girshick(2015)]{fast-rcnn47}
Ross Girshick.
\newblock Fast r-cnn.
\newblock In \emph{Proceedings of the IEEE international conference on computer vision}, pages 1440--1448, 2015.

\bibitem[Gu et~al.(2021)Gu, Lin, Kuo, and Cui]{vild9}
Xiuye Gu, Tsung-Yi Lin, Weicheng Kuo, and Yin Cui.
\newblock Open-vocabulary object detection via vision and language knowledge distillation.
\newblock \emph{arXiv preprint arXiv:2104.13921}, 2021.

\bibitem[Gupta et~al.(2019)Gupta, Dollar, and Girshick]{lvis}
Agrim Gupta, Piotr Dollar, and Ross Girshick.
\newblock Lvis: A dataset for large vocabulary instance segmentation.
\newblock In \emph{Proceedings of the IEEE/CVF conference on computer vision and pattern recognition}, pages 5356--5364, 2019.

\bibitem[Hajimiri et~al.(2025)Hajimiri, Ben~Ayed, and Dolz]{naclip15}
Sina Hajimiri, Ismail Ben~Ayed, and Jose Dolz.
\newblock Pay attention to your neighbours: Training-free open-vocabulary semantic segmentation.
\newblock In \emph{Proceedings of the Winter Conference on Applications of Computer Vision}, pages 5061--5071, 2025.

\bibitem[Huang et~al.(2025)Huang, Lu, Cun, Yu, Zhou, and Shen]{deim26}
Shihua Huang, Zhichao Lu, Xiaodong Cun, Yongjun Yu, Xiao Zhou, and Xi~Shen.
\newblock Deim: Detr with improved matching for fast convergence.
\newblock In \emph{Proceedings of the computer vision and pattern recognition conference}, pages 15162--15171, 2025.

\bibitem[Hudson and Manning(2019)]{gqa19}
Drew~A Hudson and Christopher~D Manning.
\newblock Gqa: A new dataset for real-world visual reasoning and compositional question answering.
\newblock In \emph{Proceedings of the IEEE/CVF conference on computer vision and pattern recognition}, pages 6700--6709, 2019.

\bibitem[Jia et~al.(2021)Jia, Yang, Xia, Chen, Parekh, Pham, Le, Sung, Li, and Duerig]{Jia2}
Chao Jia, Yinfei Yang, Ye~Xia, Yi-Ting Chen, Zarana Parekh, Hieu Pham, Quoc Le, Yun-Hsuan Sung, Zhen Li, and Tom Duerig.
\newblock Scaling up visual and vision-language representation learning with noisy text supervision.
\newblock In \emph{International conference on machine learning}, pages 4904--4916. PMLR, 2021.

\bibitem[Jia et~al.(2023)Jia, Yuan, He, Wu, Yu, Lin, Sun, Zhang, and Hu]{hdetr24}
Ding Jia, Yuhui Yuan, Haodi He, Xiaopei Wu, Haojun Yu, Weihong Lin, Lei Sun, Chao Zhang, and Han Hu.
\newblock Detrs with hybrid matching.
\newblock In \emph{Proceedings of the IEEE/CVF conference on computer vision and pattern recognition}, pages 19702--19712, 2023.

\bibitem[Jiang et~al.(2024)Jiang, Li, Zeng, Ren, Liu, and Zhang]{T-rex2-5}
Qing Jiang, Feng Li, Zhaoyang Zeng, Tianhe Ren, Shilong Liu, and Lei Zhang.
\newblock T-rex2: Towards generic object detection via text-visual prompt synergy.
\newblock In \emph{European Conference on Computer Vision}, pages 38--57. Springer, 2024.

\bibitem[Kamath et~al.(2021)Kamath, Singh, LeCun, Synnaeve, Misra, and Carion]{mdetr22}
Aishwarya Kamath, Mannat Singh, Yann LeCun, Gabriel Synnaeve, Ishan Misra, and Nicolas Carion.
\newblock Mdetr-modulated detection for end-to-end multi-modal understanding.
\newblock In \emph{Proceedings of the IEEE/CVF international conference on computer vision}, pages 1780--1790, 2021.

\bibitem[Lan et~al.(2024)Lan, Chen, Ke, Wang, Feng, and Zhang]{clearclip16}
Mengcheng Lan, Chaofeng Chen, Yiping Ke, Xinjiang Wang, Litong Feng, and Wayne Zhang.
\newblock Clearclip: Decomposing clip representations for dense vision-language inference.
\newblock In \emph{European Conference on Computer Vision}, pages 143--160. Springer, 2024.

\bibitem[Li et~al.(2022{\natexlab{a}})Li, Zhang, Liu, Guo, Ni, and Zhang]{dn-detr29}
Feng Li, Hao Zhang, Shilong Liu, Jian Guo, Lionel~M Ni, and Lei Zhang.
\newblock Dn-detr: Accelerate detr training by introducing query denoising.
\newblock In \emph{Proceedings of the IEEE/CVF conference on computer vision and pattern recognition}, pages 13619--13627, 2022{\natexlab{a}}.

\bibitem[Li et~al.(2022{\natexlab{b}})Li, Zhang, Zhang, Yang, Li, Zhong, Wang, Yuan, Zhang, Hwang, et~al.]{glip7}
Liunian~Harold Li, Pengchuan Zhang, Haotian Zhang, Jianwei Yang, Chunyuan Li, Yiwu Zhong, Lijuan Wang, Lu~Yuan, Lei Zhang, Jenq-Neng Hwang, et~al.
\newblock Grounded language-image pre-training.
\newblock In \emph{Proceedings of the IEEE/CVF conference on computer vision and pattern recognition}, pages 10965--10975, 2022{\natexlab{b}}.

\bibitem[Lin et~al.(2014)Lin, Maire, Belongie, Hays, Perona, Ramanan, Doll{\'a}r, and Zitnick]{COCO}
Tsung-Yi Lin, Michael Maire, Serge Belongie, James Hays, Pietro Perona, Deva Ramanan, Piotr Doll{\'a}r, and C~Lawrence Zitnick.
\newblock Microsoft coco: Common objects in context.
\newblock In \emph{European conference on computer vision}, pages 740--755. Springer, 2014.

\bibitem[Lin et~al.(2017)Lin, Goyal, Girshick, He, and Doll{\'a}r]{focal-loss31}
Tsung-Yi Lin, Priya Goyal, Ross Girshick, Kaiming He, and Piotr Doll{\'a}r.
\newblock Focal loss for dense object detection.
\newblock In \emph{Proceedings of the IEEE international conference on computer vision}, pages 2980--2988, 2017.

\bibitem[Liu et~al.(2024)Liu, Zeng, Ren, Li, Zhang, Yang, Jiang, Li, Yang, Su, et~al.]{gdino6}
Shilong Liu, Zhaoyang Zeng, Tianhe Ren, Feng Li, Hao Zhang, Jie Yang, Qing Jiang, Chunyuan Li, Jianwei Yang, Hang Su, et~al.
\newblock Grounding dino: Marrying dino with grounded pre-training for open-set object detection.
\newblock In \emph{European conference on computer vision}, pages 38--55. Springer, 2024.

\bibitem[Liu et~al.(2021)Liu, Lin, Cao, Hu, Wei, Zhang, Lin, and Guo]{swin20}
Ze~Liu, Yutong Lin, Yue Cao, Han Hu, Yixuan Wei, Zheng Zhang, Stephen Lin, and Baining Guo.
\newblock Swin transformer: Hierarchical vision transformer using shifted windows.
\newblock In \emph{Proceedings of the IEEE/CVF international conference on computer vision}, pages 10012--10022, 2021.

\bibitem[Oord et~al.(2018)Oord, Li, and Vinyals]{infonce-loss-49}
Aaron van~den Oord, Yazhe Li, and Oriol Vinyals.
\newblock Representation learning with contrastive predictive coding.
\newblock \emph{arXiv preprint arXiv:1807.03748}, 2018.

\bibitem[Plummer et~al.(2015)Plummer, Wang, Cervantes, Caicedo, Hockenmaier, and Lazebnik]{flickr30k18}
Bryan~A Plummer, Liwei Wang, Chris~M Cervantes, Juan~C Caicedo, Julia Hockenmaier, and Svetlana Lazebnik.
\newblock Flickr30k entities: Collecting region-to-phrase correspondences for richer image-to-sentence models.
\newblock In \emph{Proceedings of the IEEE international conference on computer vision}, pages 2641--2649, 2015.

\bibitem[Radford et~al.(2021)Radford, Kim, Hallacy, Ramesh, Goh, Agarwal, Sastry, Askell, Mishkin, Clark, et~al.]{Radford1}
Alec Radford, Jong~Wook Kim, Chris Hallacy, Aditya Ramesh, Gabriel Goh, Sandhini Agarwal, Girish Sastry, Amanda Askell, Pamela Mishkin, Jack Clark, et~al.
\newblock Learning transferable visual models from natural language supervision.
\newblock In \emph{International conference on machine learning}, pages 8748--8763. PmLR, 2021.

\bibitem[Redmon et~al.(2016)Redmon, Divvala, Girshick, and Farhadi]{yolo48}
Joseph Redmon, Santosh Divvala, Ross Girshick, and Ali Farhadi.
\newblock You only look once: Unified, real-time object detection.
\newblock In \emph{Proceedings of the IEEE conference on computer vision and pattern recognition}, pages 779--788, 2016.

\bibitem[Ren et~al.(2024{\natexlab{a}})Ren, Chen, Jiang, Zeng, Xiong, Liu, Ma, Shen, Gao, Jiang, et~al.]{dinox41}
Tianhe Ren, Yihao Chen, Qing Jiang, Zhaoyang Zeng, Yuda Xiong, Wenlong Liu, Zhengyu Ma, Junyi Shen, Yuan Gao, Xiaoke Jiang, et~al.
\newblock Dino-x: A unified vision model for open-world object detection and understanding.
\newblock \emph{arXiv preprint arXiv:2411.14347}, 2024{\natexlab{a}}.

\bibitem[Ren et~al.(2024{\natexlab{b}})Ren, Jiang, Liu, Zeng, Liu, Gao, Huang, Ma, Jiang, Chen, et~al.]{gdino1.5-44}
Tianhe Ren, Qing Jiang, Shilong Liu, Zhaoyang Zeng, Wenlong Liu, Han Gao, Hongjie Huang, Zhengyu Ma, Xiaoke Jiang, Yihao Chen, et~al.
\newblock Grounding dino 1.5: Advance the" edge" of open-set object detection.
\newblock \emph{arXiv preprint arXiv:2405.10300}, 2024{\natexlab{b}}.

\bibitem[Shao et~al.(2019)Shao, Li, Zhang, Peng, Yu, Zhang, Li, and Sun]{objects365}
Shuai Shao, Zeming Li, Tianyuan Zhang, Chao Peng, Gang Yu, Xiangyu Zhang, Jing Li, and Jian Sun.
\newblock Objects365: A large-scale, high-quality dataset for object detection.
\newblock In \emph{Proceedings of the IEEE/CVF international conference on computer vision}, pages 8430--8439, 2019.

\bibitem[Sim{\'e}oni et~al.(2025)Sim{\'e}oni, Vo, Seitzer, Baldassarre, Oquab, Jose, Khalidov, Szafraniec, Yi, Ramamonjisoa, et~al.]{dinov3-46}
Oriane Sim{\'e}oni, Huy~V Vo, Maximilian Seitzer, Federico Baldassarre, Maxime Oquab, Cijo Jose, Vasil Khalidov, Marc Szafraniec, Seungeun Yi, Micha{\"e}l Ramamonjisoa, et~al.
\newblock Dinov3.
\newblock \emph{arXiv preprint arXiv:2508.10104}, 2025.

\bibitem[Tian et~al.(2019)Tian, Shen, Chen, and He]{fcos32}
Zhi Tian, Chunhua Shen, Hao Chen, and Tong He.
\newblock Fcos: Fully convolutional one-stage object detection.
\newblock In \emph{Proceedings of the IEEE/CVF international conference on computer vision}, pages 9627--9636, 2019.

\bibitem[Vaswani et~al.(2017)Vaswani, Shazeer, Parmar, Uszkoreit, Jones, Gomez, Kaiser, and Polosukhin]{tfer21}
Ashish Vaswani, Noam Shazeer, Niki Parmar, Jakob Uszkoreit, Llion Jones, Aidan~N Gomez, {\L}ukasz Kaiser, and Illia Polosukhin.
\newblock Attention is all you need.
\newblock \emph{Advances in neural information processing systems}, 30, 2017.

\bibitem[Wang et~al.(2025)Wang, Liu, Chen, Lin, Han, and Ding]{yoloe4}
Ao~Wang, Lihao Liu, Hui Chen, Zijia Lin, Jungong Han, and Guiguang Ding.
\newblock Yoloe: Real-time seeing anything.
\newblock In \emph{Proceedings of the IEEE/CVF International Conference on Computer Vision}, pages 24591--24602, 2025.

\bibitem[Wang et~al.(2024)Wang, Mei, and Yuille]{sclip14}
Feng Wang, Jieru Mei, and Alan Yuille.
\newblock Sclip: Rethinking self-attention for dense vision-language inference.
\newblock In \emph{European conference on computer vision}, pages 315--332. Springer, 2024.

\bibitem[Wang et~al.(2026)Wang, Liu, Shen, Yu, He, Yu, and Chen]{ov-deim45}
Leilei Wang, Longfei Liu, Xi~Shen, Xuanlong Yu, Ying~Tiffany He, Fei~Richard Yu, and Yingyi Chen.
\newblock Ov-deim: Real-time detr-style open-vocabulary object detection with gridsynthetic augmentation.
\newblock \emph{arXiv preprint arXiv:2603.07022}, 2026.

\bibitem[Wang et~al.(2023)Wang, Liu, Du, Ding, Liao, Qi, Chen, and Liu]{oadp11}
Luting Wang, Yi~Liu, Penghui Du, Zihan Ding, Yue Liao, Qiaosong Qi, Biaolong Chen, and Si~Liu.
\newblock Object-aware distillation pyramid for open-vocabulary object detection.
\newblock In \emph{Proceedings of the IEEE/CVF conference on computer vision and pattern recognition}, pages 11186--11196, 2023.

\bibitem[Wu et~al.(2023)Wu, Zhang, Jin, Liu, and Loy]{baron8}
Size Wu, Wenwei Zhang, Sheng Jin, Wentao Liu, and Chen~Change Loy.
\newblock Aligning bag of regions for open-vocabulary object detection.
\newblock In \emph{Proceedings of the IEEE/CVF conference on computer vision and pattern recognition}, pages 15254--15264, 2023.

\bibitem[Yao et~al.(2022)Yao, Han, Wen, Liang, Xu, Zhang, Li, Xu, and Xu]{detclip40}
Lewei Yao, Jianhua Han, Youpeng Wen, Xiaodan Liang, Dan Xu, Wei Zhang, Zhenguo Li, Chunjing Xu, and Hang Xu.
\newblock Detclip: Dictionary-enriched visual-concept paralleled pre-training for open-world detection.
\newblock \emph{Advances in Neural Information Processing Systems}, 35:\penalty0 9125--9138, 2022.

\bibitem[Zhang et~al.(2022{\natexlab{a}})Zhang, Li, Liu, Zhang, Su, Zhu, Ni, and Shum]{dino17}
Hao Zhang, Feng Li, Shilong Liu, Lei Zhang, Hang Su, Jun Zhu, Lionel~M Ni, and Heung-Yeung Shum.
\newblock Dino: Detr with improved denoising anchor boxes for end-to-end object detection.
\newblock \emph{arXiv preprint arXiv:2203.03605}, 2022{\natexlab{a}}.

\bibitem[Zhang et~al.(2022{\natexlab{b}})Zhang, Zhang, Hu, Chen, Li, Dai, Wang, Yuan, Hwang, and Gao]{glipv2-43}
Haotian Zhang, Pengchuan Zhang, Xiaowei Hu, Yen-Chun Chen, Liunian Li, Xiyang Dai, Lijuan Wang, Lu~Yuan, Jenq-Neng Hwang, and Jianfeng Gao.
\newblock Glipv2: Unifying localization and vision-language understanding.
\newblock \emph{Advances in Neural Information Processing Systems}, 35:\penalty0 36067--36080, 2022{\natexlab{b}}.

\bibitem[Zhong et~al.(2022)Zhong, Yang, Zhang, Li, Codella, Li, Zhou, Dai, Yuan, Li, et~al.]{regionclip38}
Yiwu Zhong, Jianwei Yang, Pengchuan Zhang, Chunyuan Li, Noel Codella, Liunian~Harold Li, Luowei Zhou, Xiyang Dai, Lu~Yuan, Yin Li, et~al.
\newblock Regionclip: Region-based language-image pretraining.
\newblock In \emph{Proceedings of the IEEE/CVF conference on computer vision and pattern recognition}, pages 16793--16803, 2022.

\bibitem[Zhou et~al.(2022{\natexlab{a}})Zhou, Loy, and Dai]{maskclip13}
Chong Zhou, Chen~Change Loy, and Bo~Dai.
\newblock Extract free dense labels from clip.
\newblock In \emph{European conference on computer vision}, pages 696--712. Springer, 2022{\natexlab{a}}.

\bibitem[Zhou et~al.(2022{\natexlab{b}})Zhou, Girdhar, Joulin, Kr{\"a}henb{\"u}hl, and Misra]{39}
Xingyi Zhou, Rohit Girdhar, Armand Joulin, Philipp Kr{\"a}henb{\"u}hl, and Ishan Misra.
\newblock Detecting twenty-thousand classes using image-level supervision.
\newblock In \emph{European conference on computer vision}, pages 350--368. Springer, 2022{\natexlab{b}}.

\bibitem[Zhu et~al.(2020)Zhu, Su, Lu, Li, Wang, and Dai]{deformable-detr28}
Xizhou Zhu, Weijie Su, Lewei Lu, Bin Li, Xiaogang Wang, and Jifeng Dai.
\newblock Deformable detr: Deformable transformers for end-to-end object detection.
\newblock \emph{arXiv preprint arXiv:2010.04159}, 2020.

\bibitem[Zong et~al.(2023)Zong, Song, and Liu]{co-detr25}
Zhuofan Zong, Guanglu Song, and Yu~Liu.
\newblock Detrs with collaborative hybrid assignments training.
\newblock In \emph{Proceedings of the IEEE/CVF international conference on computer vision}, pages 6748--6758, 2023.

\end{thebibliography}
}

%%%%%%%%%%%%%%%%%%%%%%%%%%%%%%%%%%%%%%%%%%%%%%%%%%%%%%%%%%%%
\clearpage
\appendix

\section{Technical appendices and supplementary material}
\label{material}

\subsection{Mode implementation details}

VL-DINO is trained on eight NVIDIA L40 GPUs. The total batch sizes for VL-DINO-T and VL-DINO-L are set to 64 and 32, respectively. For the visual detection branch, we use a pretrained Swin Transformer as the backbone, and encoder and decoder are composed of six deformable encoder layers and six deformable decoder layers, respectively, with 900 object queries used for object detection. The hyperparameters are shown in Table ~\ref{tab:hyperparameters}. We adopt CLIP-B/16 as the default foundation model for visual semantic distillation, while using its corresponding text encoder to extract textual features. To reduce the training cost, we pre-compute and cache the textual features offline, allowing the text encoder to be removed from the training pipeline. The classification confidence score is calculated based on the $C$-dimensional visual-text features as follows:
\begin{equation}
\begin{gathered}
s = \sigma(\frac{\| \mathbf{v}\|_2 \|\mathbf{t}\|_2^\top}{\tau})
\end{gathered}
\label{eq:focal loss}
\end{equation}
where $\sigma(\cdot)$ denotes the sigmoid activation function, $\mathbf{v} \in \mathbb{R}^{900 \times C}$ denotes the object-query features output by the decoder, $\mathbf{t} \in \mathbb{R}^{M \times C}$ denotes $M$ text features, and $\tau$ is a learnable parameter. 

In QPSC, we generate auxiliary boxes using the same strategy as the denoising box generation mechanism in DINO. Since QPSC does not require negative samples, it uses only half the number of auxiliary queries compared with the denoising queries in DINO. We refer readers to DINO~\cite{dino17} for more implementation details of the box generation process.

In terms of the training objective, VL-DINO adopts focal loss as the classification loss $\mathcal{L}_{cls}$, $L1$ loss and GIoU loss as the box regression losses $\mathcal{L}_{box}$, $\mathcal{L}_{\mathrm{dis}}$ in Eq.~\ref{eq:dis} as the distillation loss, and bidirectional contrastive loss as alignment loss $\mathcal{L}_{\mathrm{ORSA}}$  in ORSA. Additionally, we apply auxiliary losses to both the output of encoder and each decoder layer. The overall loss is formulated as the sum of the above loss terms.

\begin{table}[!htbp]
\centering
\caption{Hyperparameters used in our experiments.}
\label{tab:hyperparameters}
\begin{tabular}{l|c}
\toprule
\textbf{Item} & \textbf{Value} \\
\midrule
optimizer & AdamW \\
lr & $1\mathrm{e}{-4}$ \\
lr\_image\_backbone & $1\mathrm{e}{-5}$ \\
weight\_decay & $0.0001$ \\
clip\_max\_norm & $0.1$ \\
encoder\_layers & $6$ \\
decoder\_layers & $6$ \\
dim\_feedforward & $2048$ \\
hidden\_dim & $256$ \\
dropout & $0.0$ \\
nheads & $8$ \\
number\_queries & $900$ \\
set\_cost\_class & $1.0$ \\
set\_cost\_bbox & $5.0$ \\
set\_cost\_giou & $2.0$ \\
ce\_loss\_coef & $2.0$ \\
bbox\_loss\_coef & $5.0$ \\
giou\_loss\_coef & $2.0$ \\
interm\_loss\_coef & $1.0$ \\
$\tau_1$ & $0.05$ \\
$\tau_2$ & $0.1$ \\
num\_label & $100$ \\
$\lambda_{\mathrm{sem}}$ & 1 \\
$\lambda_{\mathrm{rel}}$ & 1 \\
$\lambda_{\mathrm{spa}}$ & 2 \\
$\lambda_{\mathrm{ORSA}}$ & 1 \\
\bottomrule
\end{tabular}
\end{table}

\subsection{VSE details}
\begin{figure}
  \centering
  \includegraphics[width=0.6\textwidth]{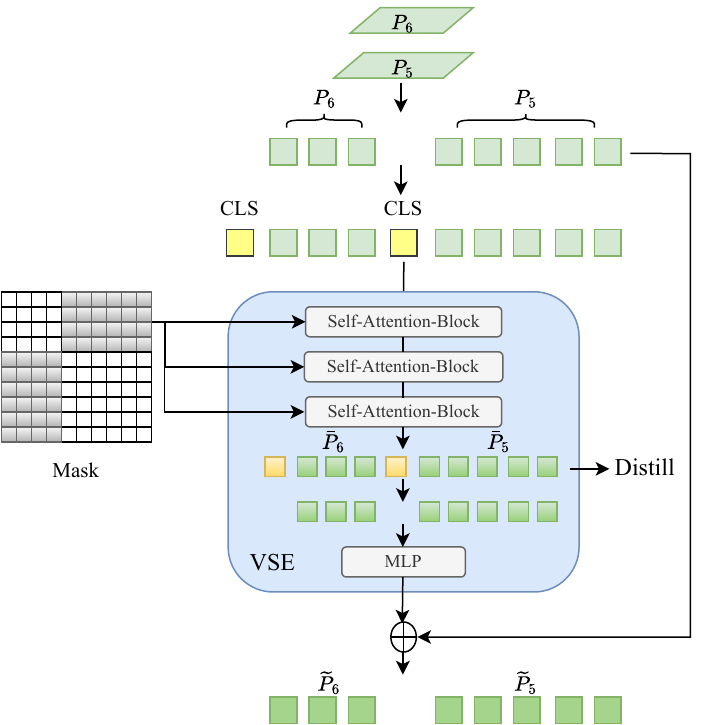}
  \caption{Details of the VSE. The attention mask prevents tokens from different feature levels from attending to each other, thereby avoiding cross-level interference between $P_5$ and $P_6$. In the mask visualization, gray regions indicate blocked attention entries, while the remaining regions allow attention computation within the same feature level. The self-attention outputs are used for distillation, while the visual tokens are further passed through an MLP and fused with the original flattened tokens to produce $\tilde{P}_5$ and $\tilde{P}_6$.
}
  \label{fig:vse}
\end{figure}

Figure ~\ref{fig:vse} shows the detailed architecture of VSE. VSE is composed of three CLIP-style self-attention blocks followed by an MLP. Given the high-level features $P_5$ and $P_6$, we first flatten them into token sequences and prepend a class embedding ($CLS$) to the tokens of each feature level. The resulting sequence is then processed by the self-attention blocks with an attention mask. The mask is designed to prevent tokens from different feature levels from attending to each other, so that the semantic modeling of $P_5$ and $P_6$ is performed independently without cross-level interference.

The outputs of the self-attention blocks are used for distillation. Afterwards, the $CLS$ tokens are discarded, and the remaining visual tokens are passed through the MLP. The resulting tokens are fused with the original flattened tokens via residual addition, yielding the enhanced token features $\tilde{P}_5$ and $\tilde{P}_6$. This design allows VSE to enhance high-level semantic representations while preserving the original token organization and avoiding undesired interference between different feature levels.

\begin{figure}[!t]
  \centering
  \includegraphics[width=0.6\textwidth]{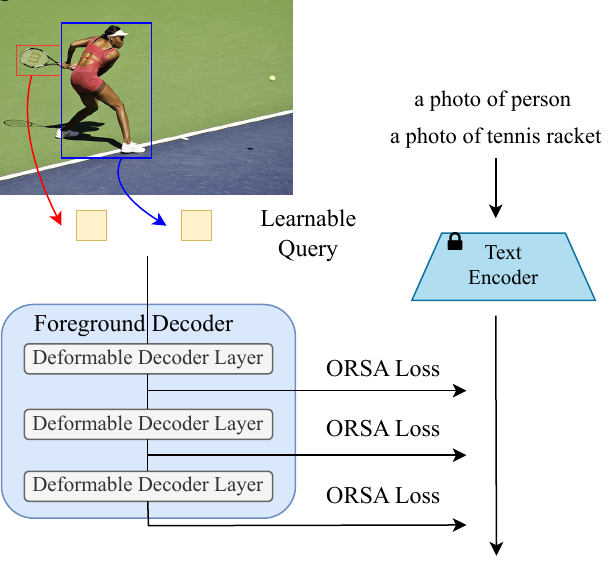}
  \caption{Framework of ORSA.
}
  \label{fig:orsa}
\end{figure}

\subsection{ORSA details}
As illustrated in Figure ~\ref{fig:orsa}, the foreground decoder in ORSA adopts the same architecture as DINO decoder, while using only three decoder layers. Different from DINO decoder, it does not include box regression. Instead, ground-truth boxes are used as reference boxes to precisely extract foreground object-region features, which are then aligned with their corresponding textual features.

Note that, for clarity of illustration, Figure ~\ref{fig:orsa} depicts ORSA as being applied to the original image. During actual training, ORSA operates on feature maps to extract object-region features, instead of directly processing the original image.

\subsection{Additional experimental results}

\subsubsection{Object query visualization}
\begin{figure}[t]
  \centering
  % --- 第一行左子图 (a) ---
  \begin{subfigure}[t]{0.48\textwidth}
    \centering
    \includegraphics[width=\linewidth]{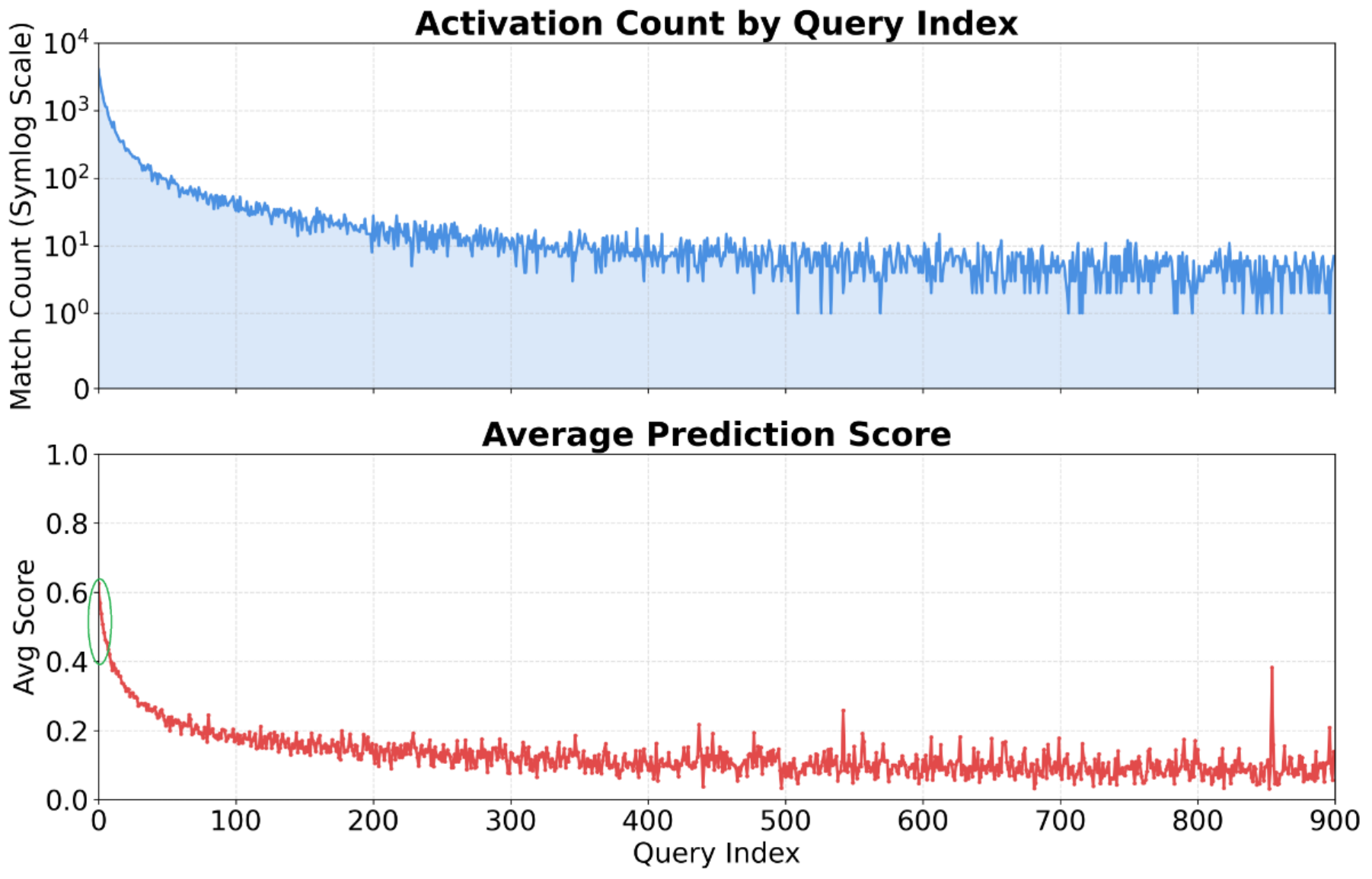}
    \caption{}
    \label{fig:query_coco_og_ori_1}
  \end{subfigure}
  \hfill
  % --- 第一行右子图 (b) ---
  \begin{subfigure}[t]{0.48\textwidth}
    \centering
    \includegraphics[width=\linewidth]{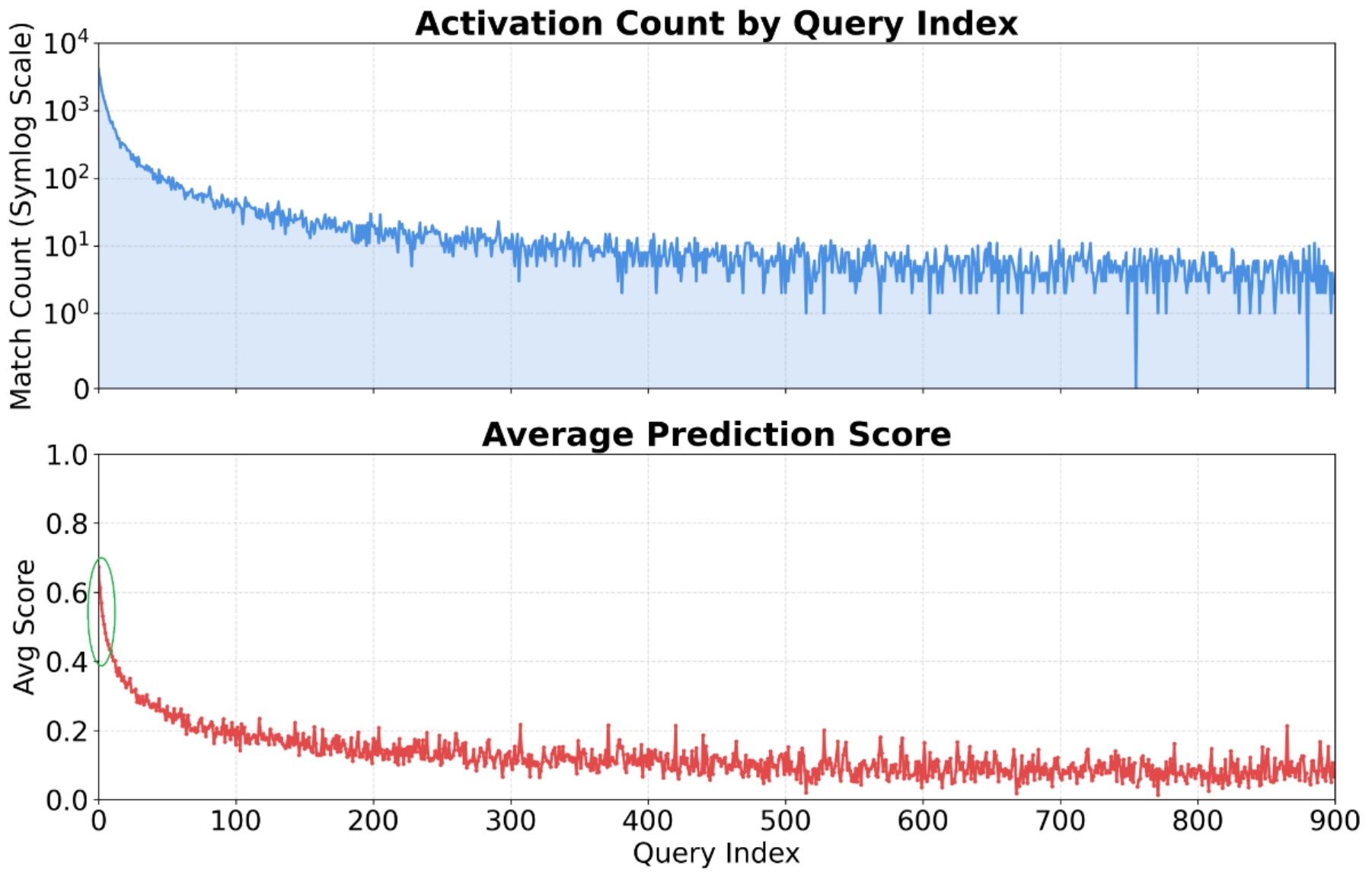}
    \caption{}
    \label{fig:query_coco_og_new_1}
  \end{subfigure}

  \vspace{0.8em}

  % --- 第二行左子图 (c) ---
  \begin{subfigure}[t]{0.48\textwidth}
    \centering
    \includegraphics[width=\linewidth]{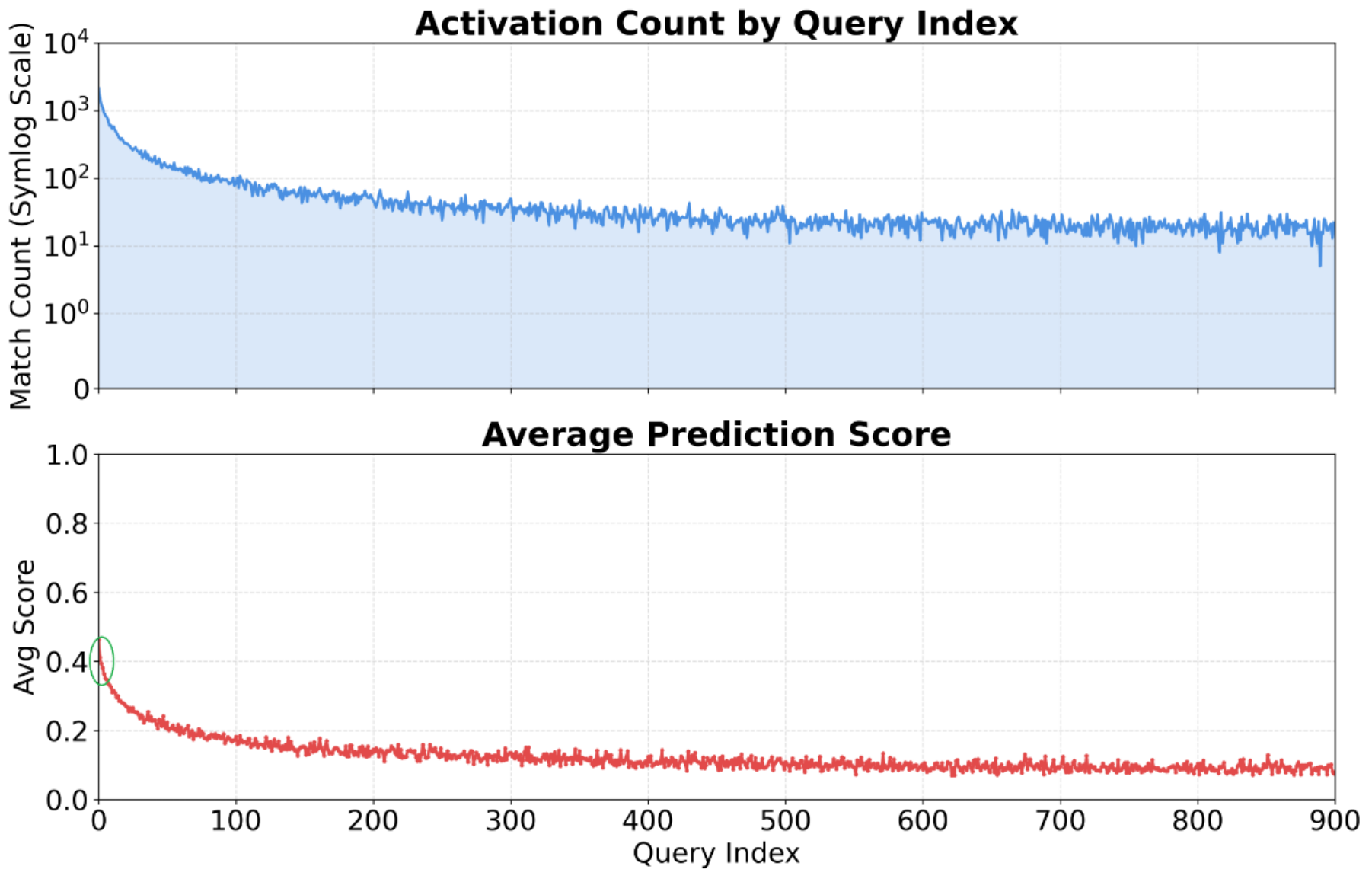}
    \caption{}
    \label{fig:query_lvis_og_ori_1}
  \end{subfigure}
  \hfill
  % --- 第二行右子图 (d) ---
  \begin{subfigure}[t]{0.48\textwidth}
    \centering
    \includegraphics[width=\linewidth]{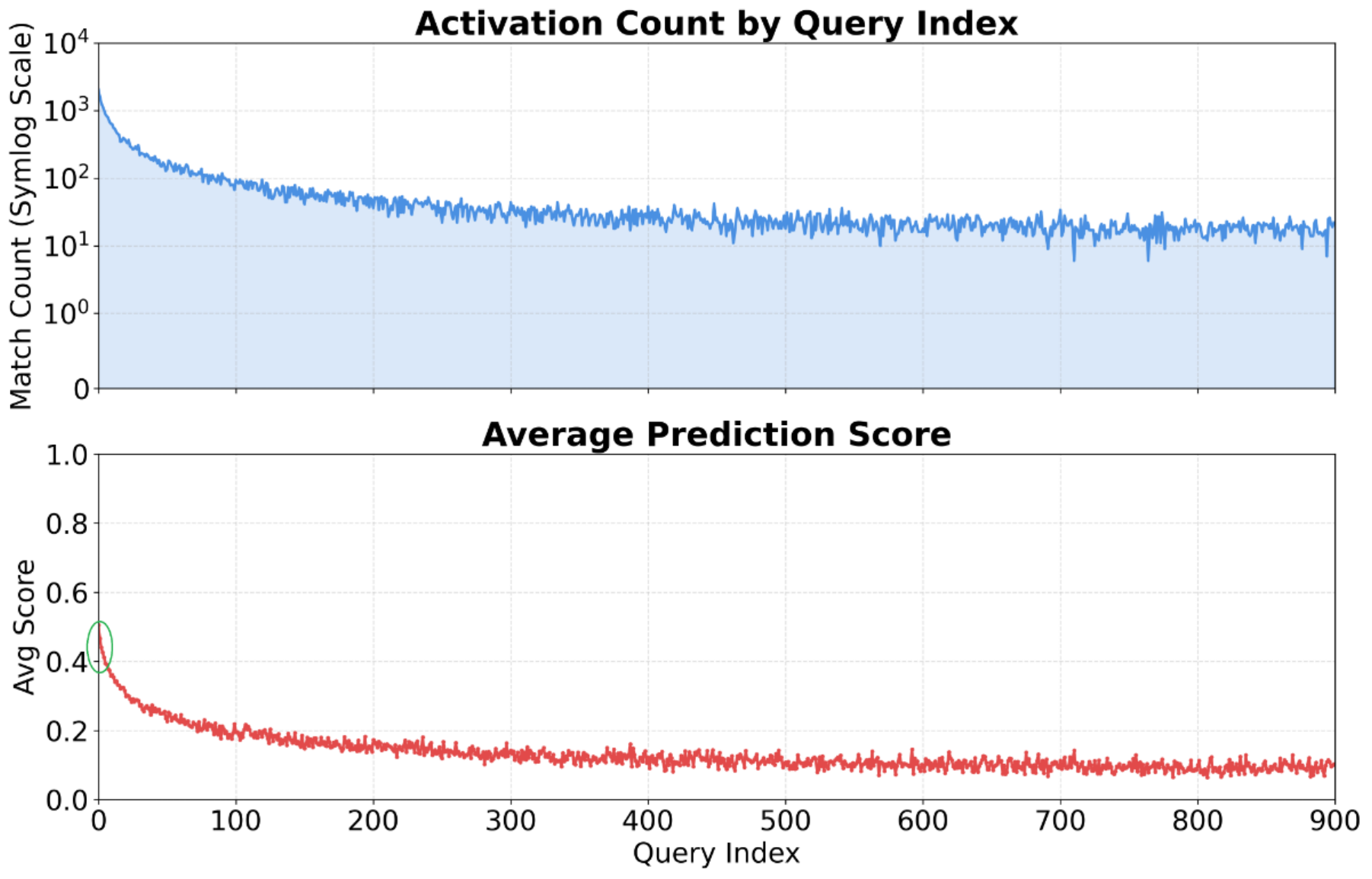}
    \caption{}
    \label{fig:query_lvis_og_new_1}
  \end{subfigure}

  \caption{Object query statistics before and after introducing QPSC on COCO and LVIS. Subfigures (a)/(b) correspond to COCO without/with QPSC, while subfigures (c)/(d) correspond to LVIS without/with QPSC.}
  \label{fig:query_visualization}
\end{figure}

In Figure ~\ref{fig:query_visualization}, we visualize the number of matches and average prediction scores of each query on the COCO and LVIS before and after introducing QPSC. The model is trained on the OG dataset until convergence. As shown in the figure, after introducing QPSC, object queries with smaller indices tend to produce higher prediction scores compared with the baseline without QPSC.

\subsubsection{Ablation study of image teacher model}
In addition to CLIP-B/16, we also use CLIP-B/32 and CLIP-L/14 as teacher models to distill visual semantic knowledge into VL-DINO-T. The experimental results are reported in Table ~\ref{table12}. CLIP-B/32 brings a smaller performance improvement than CLIP-B/16. CLIP-L/14 further improves the results, but the gain remains limited. We conjecture that this is because the features extracted by the Swin Transformer differ from those of CLIP, and such a representational gap may restrict the amount of visual semantic knowledge that can be effectively transferred through distillation.

\begin{table}[t]
    \centering
    \begin{minipage}{0.6\linewidth}
    \centering
    \caption{Ablation study of different teacher models for visual semantic distillation in VL-DINO-T.}
    \label{table12}
    \footnotesize
    \renewcommand{\arraystretch}{1.2}
    
    \begin{tabular*}{\linewidth}{@{\extracolsep{\fill}} c c c c c @{}}
      \toprule
      Teacher model & $\mathrm{AP}$ & $\mathrm{AP}_r$ & $\mathrm{AP}_c$ & $\mathrm{AP}_f$ \\
      \midrule
      CLIP-B/16 & 36.3 & 37.8 & 35.6 & 36.7 \\
      CLIP-B/32 & 35.8 & 35.8 & 34.7 & 36.8 \\
      CLIP-L/14 & 36.6 & 37.0 & 36.4 & 36.6 \\
      \bottomrule
    \end{tabular*}
    \end{minipage}
\end{table}

\subsubsection{Inference}
During both training and inference, VL-DINO can take pre-computed textual features cached offline as the classifier weights. Given an input image of size $640 \times 640$ and 1,203 textual features with a dimension of 512, we evaluate the inference performance of VL-DINO-T and VL-DINO-L on an NVIDIA L40 GPU, without including the text encoder. The corresponding performance statistics are shown in Table ~\ref{table13}.

\begin{table}[t]
    \centering
    \begin{minipage}{0.48\linewidth}
    \centering
    \caption{Inference statistics of VL-DINO-T and VL-DINO-L.}
    \label{table13}
    \footnotesize
    \renewcommand{\arraystretch}{1.2}
    
    \begin{tabular*}{\linewidth}{@{\extracolsep{\fill}} c c c c @{}}
      \toprule
      Model & Params & GFLOPS & FPS \\
      \midrule
      VL-DINO-T & 51.4M & 257G & 27.8 \\
      VL-DINO-L & 221.2M & 811G & 15.1 \\
      \bottomrule
    \end{tabular*}
    \end{minipage}
\end{table}

\subsubsection{More visualization results}
\begin{figure}[!h]
    \centering
    
    % 第一张子图
    \begin{subfigure}{0.32\textwidth}
        \centering
        % 请将 image1.png 替换为你上传的实际文件名
        \includegraphics[width=\textwidth]{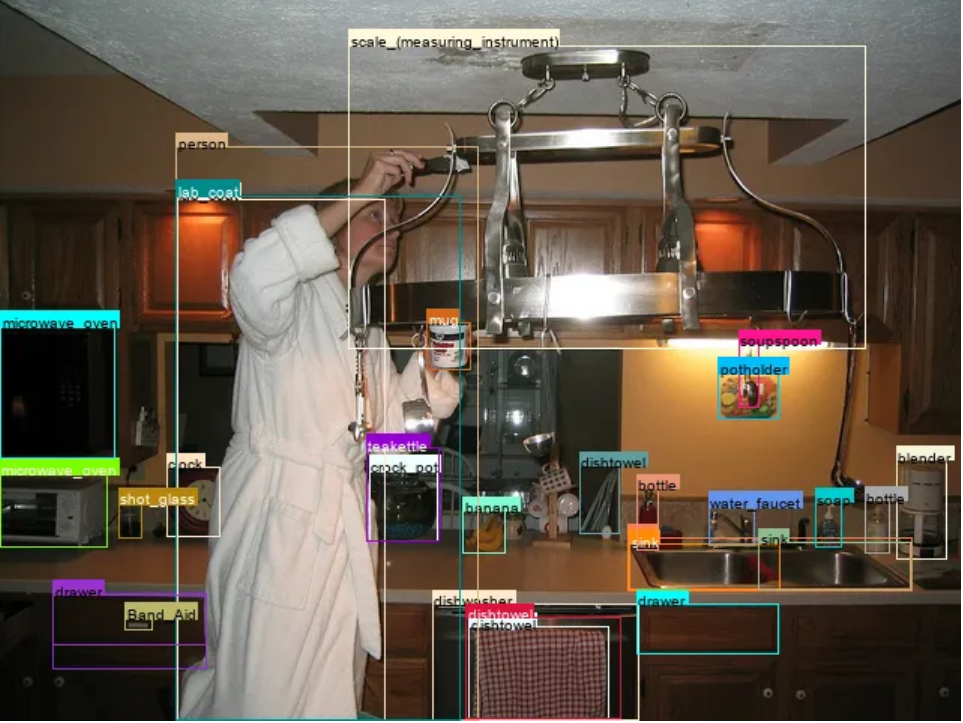}
        \caption{}
        \label{fig_sub1}
    \end{subfigure}
    \hfill % 添加弹性间距，让图片自动均匀散开
    % 第二张子图
    \begin{subfigure}{0.32\textwidth}
        \centering
        \includegraphics[width=\textwidth]{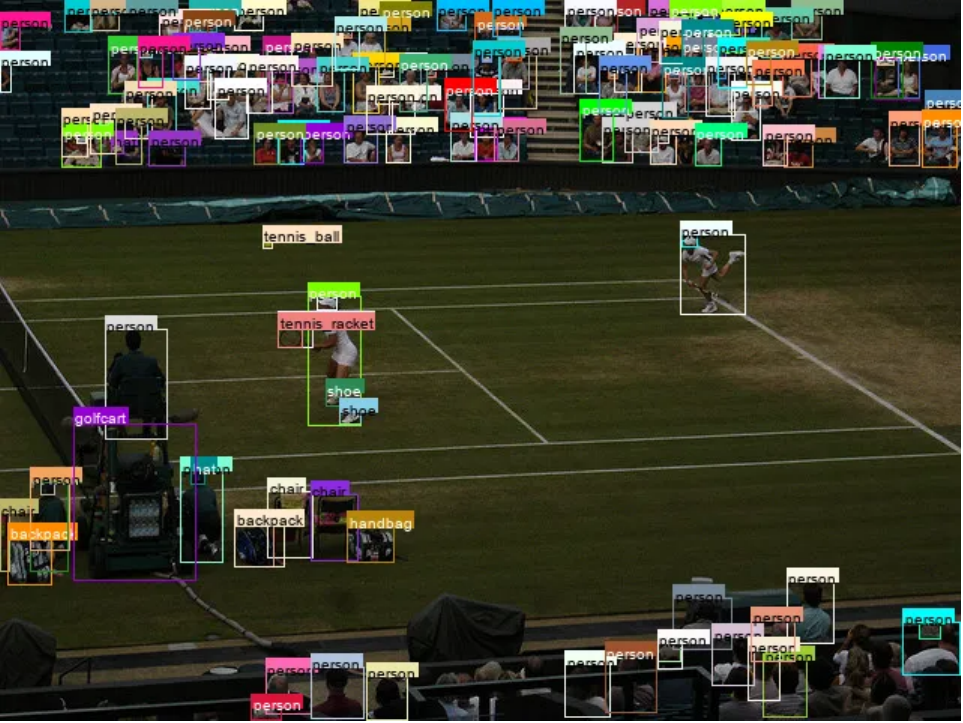}
        \caption{}
        \label{fig_sub2}
    \end{subfigure}
    \hfill
    % 第三张子图
    \begin{subfigure}{0.32\textwidth}
        \centering
        \includegraphics[width=\textwidth]{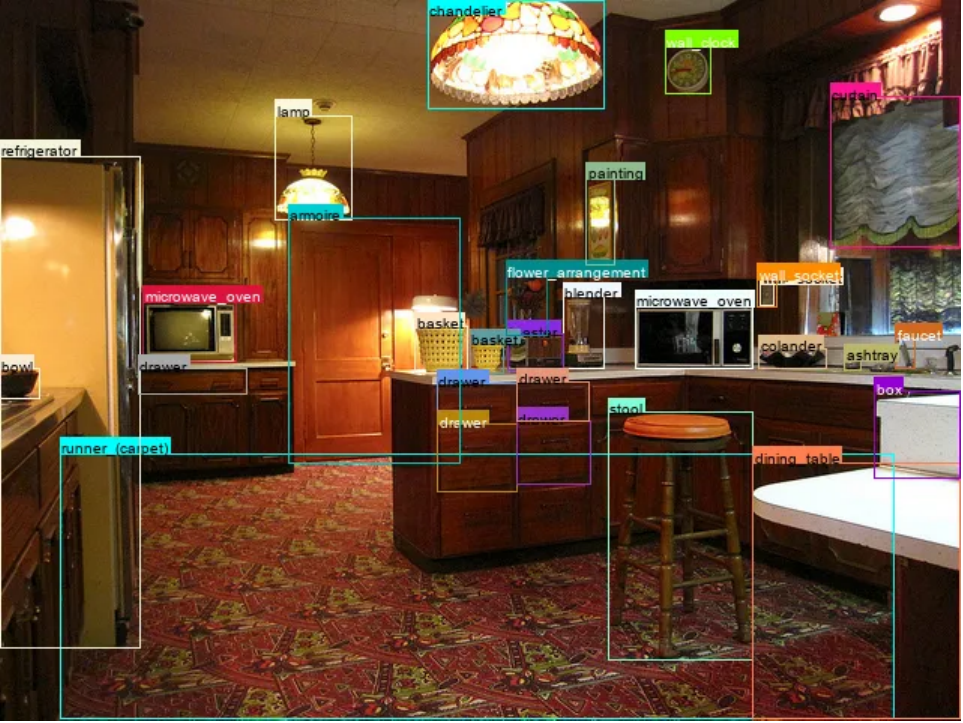}
        \caption{}
        \label{fig_sub3}
    \end{subfigure}

    \caption{Visualizations of zero-shot inference on LVIS. The pretrained VL-DINO-T model is evaluated with all 1,203 category text embeddings as input.}
    \label{fig:visualization}
    \vspace{-0.8\baselineskip}
\end{figure}

We provide additional detection visualization results on LVIS in Figure ~\ref{fig:visualization}. During inference, the model takes an image and 1,203 textual features as input.

%%%%%%%%%%%%%%%%%%%%%%%%%%%%%%%%%%%%%%%%%%%%%%%%%%%%%%%%%%%%
\clearpage
\newpage
\section*{NeurIPS Paper Checklist}

\begin{enumerate}

\item {\bf Claims}
    \item[] Question: Do the main claims made in the abstract and introduction accurately reflect the paper's contributions and scope?
    \item[] Answer: \answerYes{} % Replace by \answerYes{}, \answerNo{}, or \answerNA{}.
    \item[] Justification: We clearly state the main contributions of this work in the abstract and introduction.

\item {\bf Limitations}
    \item[] Question: Does the paper discuss the limitations of the work performed by the authors?
    \item[] Answer: \answerYes{} % Replace by \answerYes{}, \answerNo{}, or \answerNA{}.
    \item[] Justification: We discuss the limitations of our work in Section ~\ref{conclusion}.

\item {\bf Theory assumptions and proofs}
    \item[] Question: For each theoretical result, does the paper provide the full set of assumptions and a complete (and correct) proof?
    \item[] Answer: \answerNo{} % Replace by \answerYes{}, \answerNo{}, or \answerNA{}.
    \item[] Justification: This paper does not include any theoretical results.

    \item {\bf Experimental result reproducibility}
    \item[] Question: Does the paper fully disclose all the information needed to reproduce the main experimental results of the paper to the extent that it affects the main claims and/or conclusions of the paper (regardless of whether the code and data are provided or not)?
    \item[] Answer: \answerYes{} % Replace by \answerYes{}, \answerNo{}, or \answerNA{}.
    \item[] Justification: We specify the experimental configurations in the main paper and report the detailed hyperparameter settings in the supplementary material.

\item {\bf Open access to data and code}
    \item[] Question: Does the paper provide open access to the data and code, with sufficient instructions to faithfully reproduce the main experimental results, as described in supplemental material?
    \item[] Answer: \answerNo{} % Replace by \answerYes{}, \answerNo{}, or \answerNA{}.
    \item[] Justification: All datasets used in this work are publicly available, and the code will be made open-source upon paper acceptance.

\item {\bf Experimental setting/details}
    \item[] Question: Does the paper specify all the training and test details (e.g., data splits, hyperparameters, how they were chosen, type of optimizer) necessary to understand the results?
    \item[] Answer: \answerYes{} % Replace by \answerYes{}, \answerNo{}, or \answerNA{}.
    \item[] Justification: We include the training and testing details in Section~\ref{experiment} and the supplementary material.

\item {\bf Experiment statistical significance}
    \item[] Question: Does the paper report error bars suitably and correctly defined or other appropriate information about the statistical significance of the experiments?
    \item[] Answer: \answerNo{} % Replace by \answerYes{}, \answerNo{}, or \answerNA{}.
    \item[] Justification: All experiments are conducted with the same fixed random seed to ensure reproducibility. Due to the high computational cost of our experiments, we do not report error bars.

\item {\bf Experiments compute resources}
    \item[] Question: For each experiment, does the paper provide sufficient information on the computer resources (type of compute workers, memory, time of execution) needed to reproduce the experiments?
    \item[] Answer: \answerYes{} % Replace by \answerYes{}, \answerNo{}, or \answerNA{}.
    \item[] Justification: We describe the computational resources used in our experiments in Section ~\ref{implementation details} and the supplementary material.
    
\item {\bf Code of ethics}
    \item[] Question: Does the research conducted in the paper conform, in every respect, with the NeurIPS Code of Ethics \url{https://neurips.cc/public/EthicsGuidelines}?
    \item[] Answer: \answerYes{} % Replace by \answerYes{}, \answerNo{}, or \answerNA{}.
    \item[] Justification: All research conducted in this paper complies with the NeurIPS Code of Ethics.

\item {\bf Broader impacts}
    \item[] Question: Does the paper discuss both potential positive societal impacts and negative societal impacts of the work performed?
    \item[] Answer: \answerYes{} % Replace by \answerYes{}, \answerNo{}, or \answerNA{}.
    \item[] Justification:  We provide a discussion of broader impacts in Section ~\ref{conclusion}.
    
\item {\bf Safeguards}
    \item[] Question: Does the paper describe safeguards that have been put in place for responsible release of data or models that have a high risk for misuse (e.g., pre-trained language models, image generators, or scraped datasets)?
    \item[] Answer: \answerNA{} % Replace by \answerYes{}, \answerNo{}, or \answerNA{}.
    \item[] Justification: The proposed model will not pose the aforementioned risks.

\item {\bf Licenses for existing assets}
    \item[] Question: Are the creators or original owners of assets (e.g., code, data, models), used in the paper, properly credited and are the license and terms of use explicitly mentioned and properly respected?
    \item[] Answer: \answerYes{} % Replace by \answerYes{}, \answerNo{}, or \answerNA{}.
    \item[] Justification: We cite all models, datasets, and code used in this work, and comply with their licenses.

\item {\bf New assets}
    \item[] Question: Are new assets introduced in the paper well documented and is the documentation provided alongside the assets?
    \item[] Answer: \answerNA{} % Replace by \answerYes{}, \answerNo{}, or \answerNA{}.
    \item[] Justification: This paper does not release new assets.

\item {\bf Crowdsourcing and research with human subjects}
    \item[] Question: For crowdsourcing experiments and research with human subjects, does the paper include the full text of instructions given to participants and screenshots, if applicable, as well as details about compensation (if any)? 
    \item[] Answer: \answerNA{} % Replace by \answerYes{}, \answerNo{}, or \answerNA{}.
    \item[] Justification: This paper does not involve crowdsourcing or research with human subjects.

\item {\bf Institutional review board (IRB) approvals or equivalent for research with human subjects}
    \item[] Question: Does the paper describe potential risks incurred by study participants, whether such risks were disclosed to the subjects, and whether Institutional Review Board (IRB) approvals (or an equivalent approval/review based on the requirements of your country or institution) were obtained?
    \item[] Answer: \answerNA{} % Replace by \answerYes{}, \answerNo{}, or \answerNA{}.
    \item[] Justification: This paper does not involve crowdsourcing or research with human subjects.

\item {\bf Declaration of LLM usage}
    \item[] Question: Does the paper describe the usage of LLMs if it is an important, original, or non-standard component of the core methods in this research? Note that if the LLM is used only for writing, editing, or formatting purposes and does \emph{not} impact the core methodology, scientific rigor, or originality of the research, declaration is not required.
    %this research? 
    \item[] Answer: \answerNA{} % Replace by \answerYes{}, \answerNo{}, or \answerNA{}.
    \item[] Justification: Large language models were used solely for writing and editing assistance, and were not used for any other research purposes.

\end{enumerate}

\end{document}